\documentclass[10pt,conference,a4paper]{IEEEtran}
\usepackage{graphicx}
\usepackage{amsmath,amssymb} 
\usepackage{color}
\usepackage{subfigure}
\usepackage{epstopdf}
\usepackage{epsfig}
\newcommand{\bftab}{\fontseries{b}\selectfont}
\ifCLASSINFOpdf
\else
\fi

\begin{document}
%
\title{Visualization of Hyperspectral Images Using Moving Least Squares}

\author
{\IEEEauthorblockN{Danping Liao}
\IEEEauthorblockA{College of Computer Science\\Zhejiang University, HangZhou, China\\
liaodanping@gmail.com}
\and
\IEEEauthorblockN{Siyu Chen}
\IEEEauthorblockA{College of Computer Science\\Zhejiang University, HangZhou, China\\
sychen@zju.edu.cn}
\and
\IEEEauthorblockN{Yuntao Qian}
\IEEEauthorblockA{College of Computer Science\\Zhejiang University, HangZhou, China\\
ytqian@zju.edu.cn}
}

\maketitle

\begin{abstract}
Displaying the large number of bands in a hyperspectral image (HSI) on a trichromatic monitor has been an active research topic.
The visualized image shall convey as much information as possible
from the original data and facilitate image interpretation.
Most existing methods display HSIs in false colors, which contradict with human's experience and expectation.
In this paper, we propose a nonlinear approach to visualize an input HSI with natural colors by taking advantage of a corresponding RGB image.
Our approach is based on Moving Least Squares (MLS), an interpolation scheme for reconstructing a surface from a set of control points, which in our case is a set of matching pixels between the HSI and the corresponding RGB image.
Based on MLS, the proposed method solves for each spectral signature a unique transformation so that the nonlinear structure of the HSI can be preserved.
The matching pixels between a pair of HSI and RGB image can be reused to display other HSIs captured by the same imaging sensor with natural colors.
Experiments show that the output images of the proposed method not only have natural colors but also maintain the visual information necessary for human analysis.
\end{abstract}

\section{Introduction}
The high spectral resolution of a hyperspectral image (HSI) enables accurate target detection and classification.
Human interaction with HSIs is crucial for image interpretation and analysis.
Displaying an (HSI) is a challenging task since an HSI contains much more bands than the capability of a trichromatic monitor.
One common solution is to consider HSI visualization as a dimension reduction problem where the number of bands is reduced to 3 for a color visualization.
Although the requirements of HSI visualization are task dependent, there are some common goals such as information preservation, consistent rendering and natural palette~\cite{jacobson2005design}.

A straightforward way for color representation is to select three of the original bands and map them to the RGB space~\cite{le2011constrained, su2014hyperspectral}.
However, band selection methods only take the selected spectrum into account.
As a result, the information in other bands is ignored.
To preserve the information across the full wavelength range, dimension reduction methods such as independent component analysis (ICA)~\cite{zhu2007evaluation} and  principal component analysis (PCA)~\cite{tsagaris2005fusion} have been applied to HSI visualization.
These linear transformation models are based on the global distribution of data, thus the intrinsic characteristics of HSIs such as the nonlinear and local structures are not well preserved.

To preserve the nonlinear structure of HSIs, manifold learning methods such as isometric feature mapping (ISOMAP)~\cite{bachmann2005exploiting} and locally linear embedding (LLE)~\cite{crawford2011exploring} have been applied.
To preserve the pairwise spectral distance in HSIs, Mignotte proposed a nonstationary Markov random field model~\cite{mignotte2010multiresolution}
and later extended this approach to address the tradeoff between
the preservation of spectral distances and the separability of
features~\cite{mignotte2012bicriteria}.
Kotwal and Chaudhuria used a nonlinear bilateral filter with edge preserving characteristic to compute the weights of bands at each pixel for band image fusion~\cite{kotwal2010visualization}.
These nonlinear approaches demonstrate excellent performance in preserving the intrinsic information of HSIs.

While most existing methods try to convey as much information as possible from the original data to the visualized image,
most of them display HSIs in false colors.
In general false color visualizations can be hard to interpret when object colors are very different from their natural appearance.
Moreover, most data-adaptive methods suffer the problem of ``inconsistent rendering'', i.e., the same objects/materials being displayed with very different colors in different images, which also hinders the interpretation of HSIs.
Therefore, ``natural palette'' and ``consistent rendering'' gradually become two important criteria for evaluating the quality of HSI visualization.

To generate natural looking visualizations, Jacobson \emph{et al}.~\cite{jacobson2005design} proposed a linear spectral weighting function
to fuse all the bands in an HSI. The weighting function is a stretched version of the
CIE 1964 tristimulus color matching functions in the visible range.
As the weighting function is fixed, this model can generate consistent colors. However, compared with other data adaptive methods, the model preserves less  information for each specific images.
Another approach for displaying HSIs with natural colors was proposed by Connah \emph{et al}.~\cite{connah2014spectral}, which takes advantage of a corresponding RGB image.
To obtain a natural-looking visualization, the gradient of each pixel in the visualization is constrained to be the same with that of its matching pixel in the corresponding RGB image.
This method requires pixel-wise matching between the HSI and the corresponding RGB image, which is sometimes difficult to achieve especially when the HSI and the RGB image are acquired by sensors mounted on different platforms, such as the airplane and satellite, with different geometrical distortions.
Liao \emph{et al}.~\cite{liaomanifold} proposed an approach to display HSIs with natural colors based on manifold alignment between an HSI and a high resolution RGB image.
The HSI and the RGB image are firstly projected to a common space by manifold alignment,
Then the HSI is projected from the common space to the RGB space to generate a final image that has similar colors to the corresponding RGB image.
One advantage of the approach is that it only requires a small number of matching pixels.
However, the estimated mapping function from the spectral space to RGB space is linear, which shows some limitation in representing the local structure of the HSI.

In this paper, we propose a nonlinear method to display HSIs with natural colors.
Our goal is to produce an output image which not only maintains the information of the input HSI but also has similar colors to a corresponding RGB image.
Our method is based on Moving Least Squares (MLS), which is a local and nonlinear interpolation scheme for reconstructing a surface from a set of control points~\cite{lancaster1981surfaces}.
In the task of HSI visualization, we use MLS to generate a color representation of an input HSI through a set of matching pixels between the HSI and the corresponding RGB image.
Compared with the methods that use a single transformation to map all the pixels in
the HSI to the RGB space, the proposed method estimates a transformation for each pixel
individually by solving a unique weighted least squares problem for each spectral signature.
The strong locality of MLS allows our model to preserve the local structure of HSIs very well.

One advantage of the proposed method is that the matching pixels between a pair of HSI and RGB image can be reused to visualize other similar HSIs acquired by the same imaging sensor.
Since the matching pixels set up the mapping relation between the high dimensional spectral space and the RGB space, the same/similar objects in different images are presented with consistent colors.
Another advantage is that our model does not require precise pixel-wise matching between the HSI and RGB image.
Instead, it only requires a small number of matching pairs that represent the same/similar objects, which greatly increases the flexibility of the approach.

The rest of the paper is organized as follows.
Section~\ref{movingLeastSquares} gives a brief description of MLS and some related works.
Section~\ref{Hyperspetral Image Visualization based on MLS} presents the MLS based HSI visualization approach. Some application scenarios are also discussed in this section. The experimental results are described in Section~\ref{Experiments}.
Finally, conclusions are drawn in Section~\ref{Conclusions}.
\section{Moving Least Squares}
\label{movingLeastSquares}
Moving Least Squares (MLS) is a scattered point interpolation technique for generating surfaces~\cite{lancaster1981surfaces}.
Within the framework of MLS, a continuous function can be reconstructed from a set of point samples (control points).
The technique is attractive since the surface is reconstructed
by local computations and it generates a surface that is smooth
everywhere.

Based on the distances between the control points and an input point,
MLS solves a unique weighted least squares to evaluate each input point individually.
Let $\textbf{X}$ be a set of control points and $\textbf{Y}$ the given function values of the control points.
Given an arbitrary point $\textbf{x}$ to be evaluated, MLS aims to find a transformation $f_x$ to map it to the target space.
The optimal function $f_x$ is acquired by minimizing the following weighted least squares
\begin{equation}
\label{MLSobj}
    \sum_{k}w_k\mid f_x(\textbf{X}_k)-\textbf{Y}_k\mid^2
\end{equation}
where $\textbf{X}_k$ is the $k$th control point and $\textbf{Y}_k$ is its given function value.
The summation is taken over all the control points.
The weights $w_k$ have the form
\begin{equation}
w_k=\frac{1}{|\textbf{x}-\textbf{X}_k|^{2\alpha}},
\end{equation}
which are larger for the control points that are closer to $\textbf{x}$.
The name \emph{Moving Least Squares} comes from the fact that the weights $w_k$ change depending on the point $\textbf{x}$ to be evaluated.
Therefore, MLS obtains a different transformation $f_x$ for different $\textbf{x}$.

The MLS approach has been successfully applied in image deformation~\cite{schaefer2006image},
surface reconstruction~\cite{fleishman2005robust}, and image superresolution and denoising~\cite{bose2006superresolution}.
Recently, Hwang \emph{et al}.~\cite{hwang2014color} used MLS to transfer the colors from a source RGB image to a target RGB image.
The matching pixels between the input images serve as the control points for interpolation.
For each RGB value in the target image, the method computes an affine function to map it to a desired color.
The goal for estimating the optimal affine function for each color is formalized as minimizing the summation of a set of weighted least squares over the control points.

In this paper, we apply MLS to display HSIs in natural colors.
Our work shares the same idea with~\cite{hwang2014color} in that we try to ``transfer'' the colors of a corresponding RGB image to the visualization of an HSI.
The main contribution of this work is that we further formalize the objective function
in Eq.(\ref{MLSobj}) in matrix form, and derive a solution in the form of matrix
multiplication, which can be easily parallelized and be computed more efficiently
compared to the solutions in summation form.
\section{Hyperspetral Image Visualization Based on MLS}
\label{Hyperspetral Image Visualization based on MLS}
We introduce an MLS based mechanism to display an HSI with natural colors.
The key idea is to generate an output 3D image through a set of matching pixels between the HSI and a corresponding RGB image.
In this section we first introduce the MLS algorithm for HSI visualization, followed by a coarse-to-fine searching process to find matching pixels.
Finally, two other visualization scenarios are discussed.
\subsection{Moving Least Squares Framework}
Let $\textbf{U}\in R^{p\times n}$ and $\textbf{V}\in R^{3\times n}$ be a set of matching pixel values from an HSI and an RGB image respectively, where $p$ is the number of bands in the HSI, and $n$ is the number of matching pixel pairs.
These matching pixels will serve as the control points to construct the output color image.
Let $\textbf{U}_k$ and $\textbf{V}_k$ denote the $k$th pair of matching pixels.
Given a high dimensional spectral signature $\textbf{x}$ in the HSI, we aim to find a transformation $f_x$ to transform it to the 3D RGB space.
Within the framework of MLS, the goal is to find the $f_x$ that minimizes the following weighted least squares
\begin{equation}
\label{obj}
    G(f_x)=\sum_{k=1}^{n}w_k\mid f_x(\textbf{U}_k)-\textbf{V}_k\mid^2.
\end{equation}
As the spectral angle distance (SAD) is commonly used to measure the distance between a pair of pixels in an HSI, the weight $w_k$ for an input $\textbf{x}$ is defined as
\begin{equation}
\label{w}
w_k=\frac{1}{SAD(\textbf{x},\textbf{U}_k)}
\end{equation}
where
\begin{equation}
SAD(\textbf{x},\textbf{U}_k)=\arccos(\frac{\textbf{x}\cdot \textbf{U}_k}{\|\textbf{x}\|\|\textbf{U}_k\|}).
\end{equation}
Note that weights $w_k$ change depending on the spectral signature $\textbf{x}$ to be evaluated,
thus the proposed method solves a different weighted least problem for different $\textbf{x}$.
Therefore, the optimal transformation $f_x$ varies for each $\textbf{x}$,
which allows our method to preserve the local and nonlinear structure of HSIs.

In the image deformation work~\cite{schaefer2006image}, a rigid transformation $f_x$ was chosen since it maintains the geometric properties with less degrees of freedom. In the color transfer task~\cite{hwang2014color}, $f_x$ was set to an affine transformation to model different elements of the color deformation.
In this work, we also choose affine transformation to model the nonlinearities in HSIs.
Thus $f_x$ is represented by
\begin{equation}
\label{f}
f_x(\textbf{x})=\textbf{F}_x^T\textbf{x}+\textbf{b}_x
\end{equation}
where $\textbf{F}_x$ is a $p\times 3$ matrix representing a linear transformation and $\textbf{b}_x$ is a $3\times 1$ vector representing a translation.
Taking the partial derivative of Eq.(\ref{obj}) with respect to $\textbf{b}_x$, we get
\begin{equation}
\textbf{b}_x=\overline{\textbf{v}}-\textbf{F}_x^T{\overline{\textbf{u}}}
\end{equation}
with $\overline{\textbf{u}}$ and $\overline{\textbf{v}}$ being the weighted centroids:
\begin{equation}
\overline{\textbf{u}}=\frac{\sum_k{w_k}\textbf{U}_k}{\sum_k{w_k}},~~\overline{\textbf{v}}=\frac{\sum_k{w_k}\textbf{V}_k}{\sum_k{w_k}}.
\end{equation}
Eq.(\ref{f}) can then be rewritten as
\begin{equation}
f_x=\textbf{F}_x^T(\textbf{x}-\overline{\textbf{u}})+\overline{\textbf{v}}
\end{equation}
and Eq.(\ref{obj}) becomes
\begin{equation}
\label{newobj}
G(\textbf{F}_x)=\sum_{k=1}^{n}w_k|\textbf{F}_x^T\overline{\textbf{U}}_k-\overline{\textbf{V}}_k|^2
\end{equation}
where  $\overline{\textbf{U}}$ and $\overline{\textbf{V}}$ are acquired by subtracting $\overline{\textbf{u}}$ and $\overline{\textbf{v}}$ from each column of $\textbf{U}$ and $\textbf{V}$ respectively.
Eq.(\ref{newobj}) can be rewritten in matrix form as
\begin{equation}
\label{matrixobj}
G(\textbf{F}_x)=tr(\textbf{F}_x^T\overline{\textbf{U}}\textbf{W}\overline{\textbf{U}}^T\textbf{F}_x
-2\textbf{F}_x^T\overline{\textbf{U}}\textbf{W}\overline{\textbf{V}}+\overline{\textbf{V}}\textbf{W}\overline{\textbf{V}}^T)
\end{equation}
where $\textbf{W}$ is an $n\times n$ diagonal matrix with $\textbf{W}_{kk}=w_k$, and $tr(\cdot)$ denotes the trace of a matrix.
This objective function is convex and differentiable. Its derivative with respect to $\textbf{F}_x$ is
\begin{equation}
\label{derivative}
\frac{\partial G}{\partial \textbf{F}_x}=2\overline{\textbf{U}}\textbf{W}\overline{\textbf{U}}^T\textbf{F}_x-2\overline{\textbf{U}}\textbf{W}\overline{\textbf{V}}.
\end{equation}
The optimal $\textbf{F}_x$ that minimizes the objective function can be obtained by
\begin{equation}
\label{solution}
\textbf{F}_x=(\overline{\textbf{U}}\textbf{W}\overline{\textbf{U}}^T)^{-1}\overline{\textbf{U}}\textbf{W}\overline{\textbf{V}}^T.
\end{equation}
When $\textbf{F}_x$ and $\textbf{b}_x$ are acquired, a spectral signature $\textbf{x}$ is transformed to an RGB color by Eq.(\ref{f}).
A natural-looking representation of the HSI can be obtained after mapping all the pixels in the HSI to the RGB space.
\subsection{Finding the Matching Pixels Between an HSI and an RGB Image}
\label{SectionMatching}
\begin{figure}[t]
\centering
\subfigure[The 50th band of the HSI]
{\includegraphics[angle=90,width=0.7\linewidth]{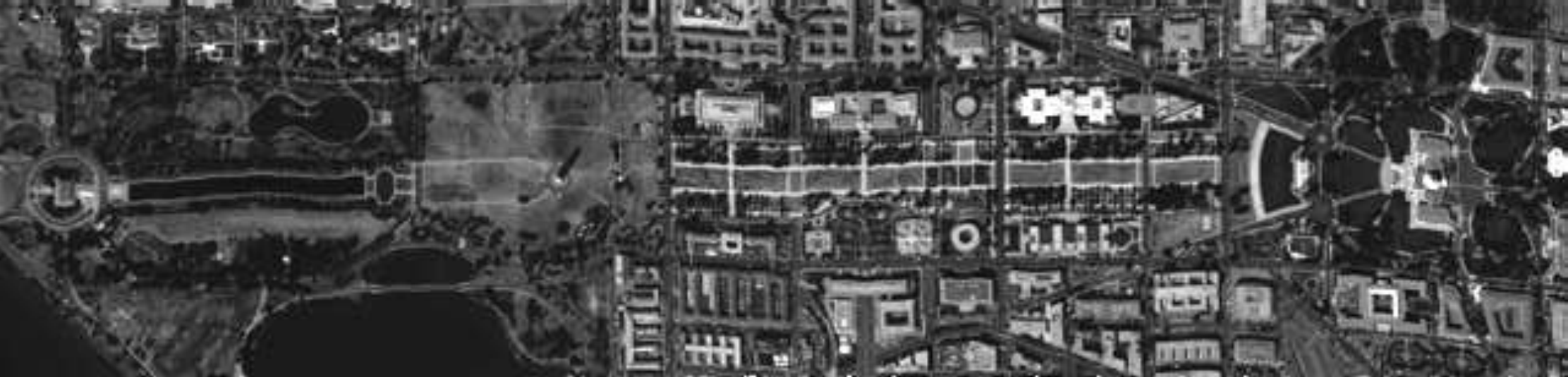}
\label{fig:band50}
}
\subfigure[The corresponding RGB image]
{\includegraphics[angle=90,width=0.7\linewidth]{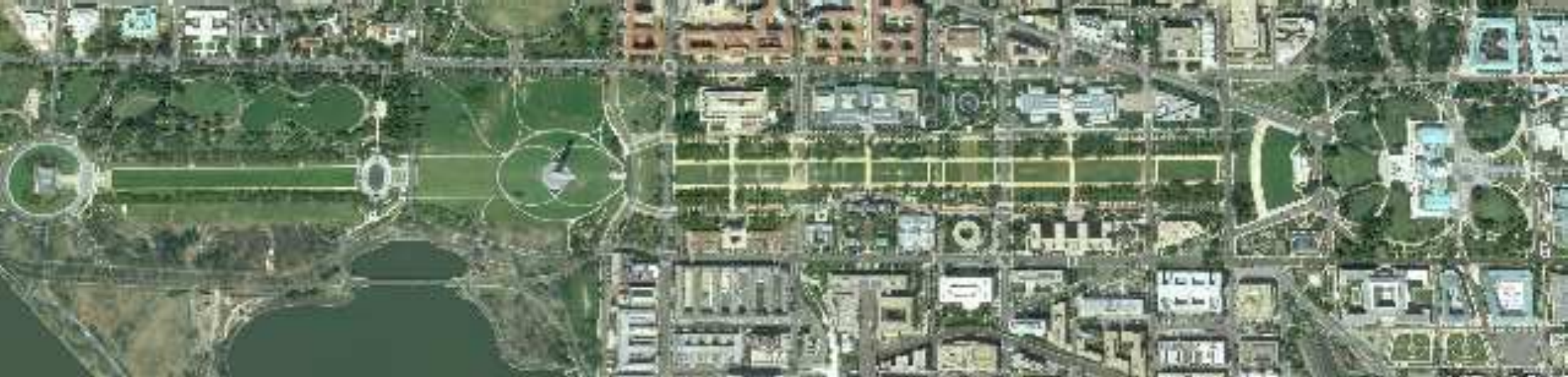}
\label{fig:registered}
}
\caption{An HSI and its corresponding RGB image of the Washington D.C. mall.}\label{DC}
\end{figure}
\begin{figure}
\centering
\subfigure[]
{\includegraphics[width=0.3\linewidth]{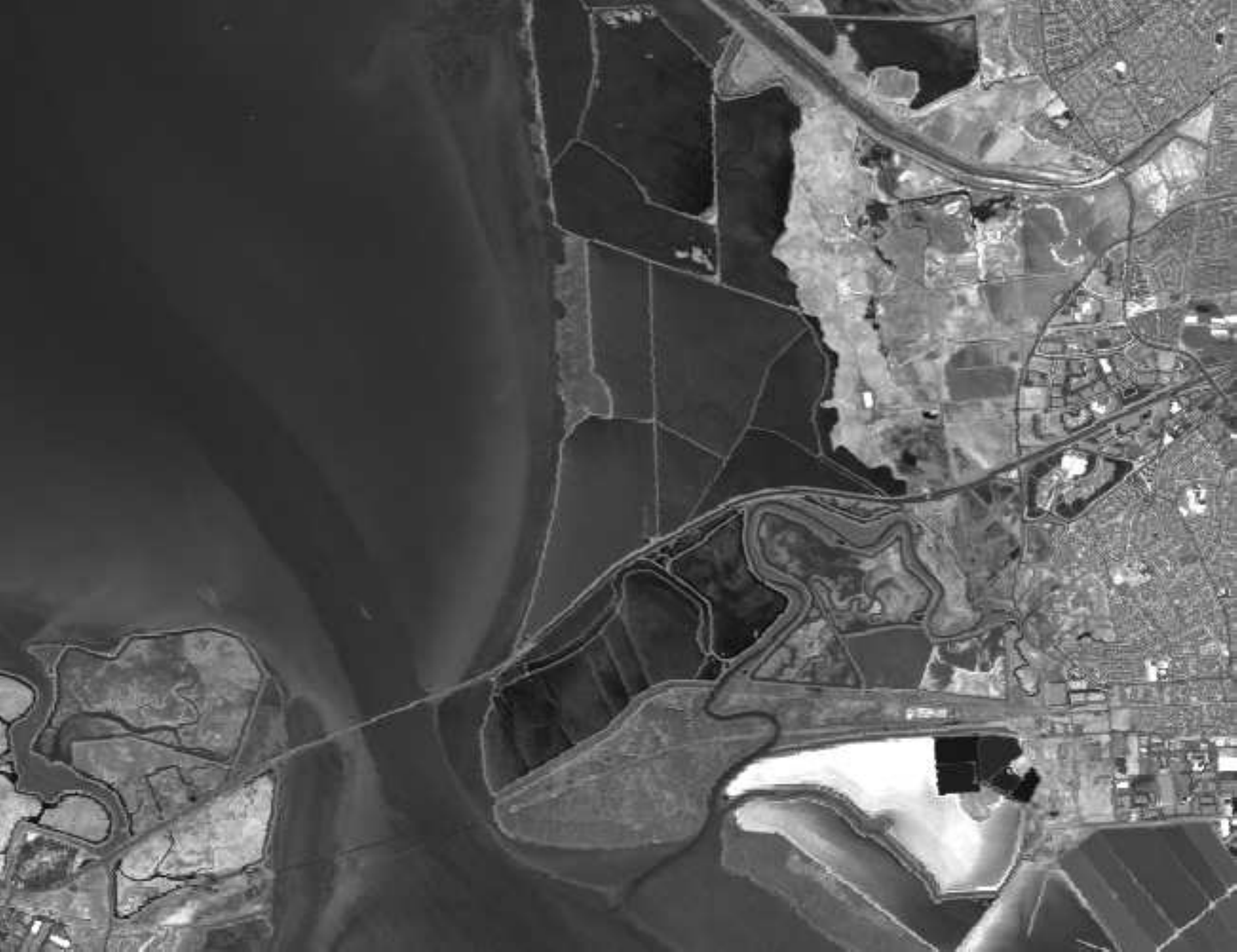}
\label{fig:MoffettBand50}
}
\subfigure[]
{\includegraphics[width=0.3\linewidth]{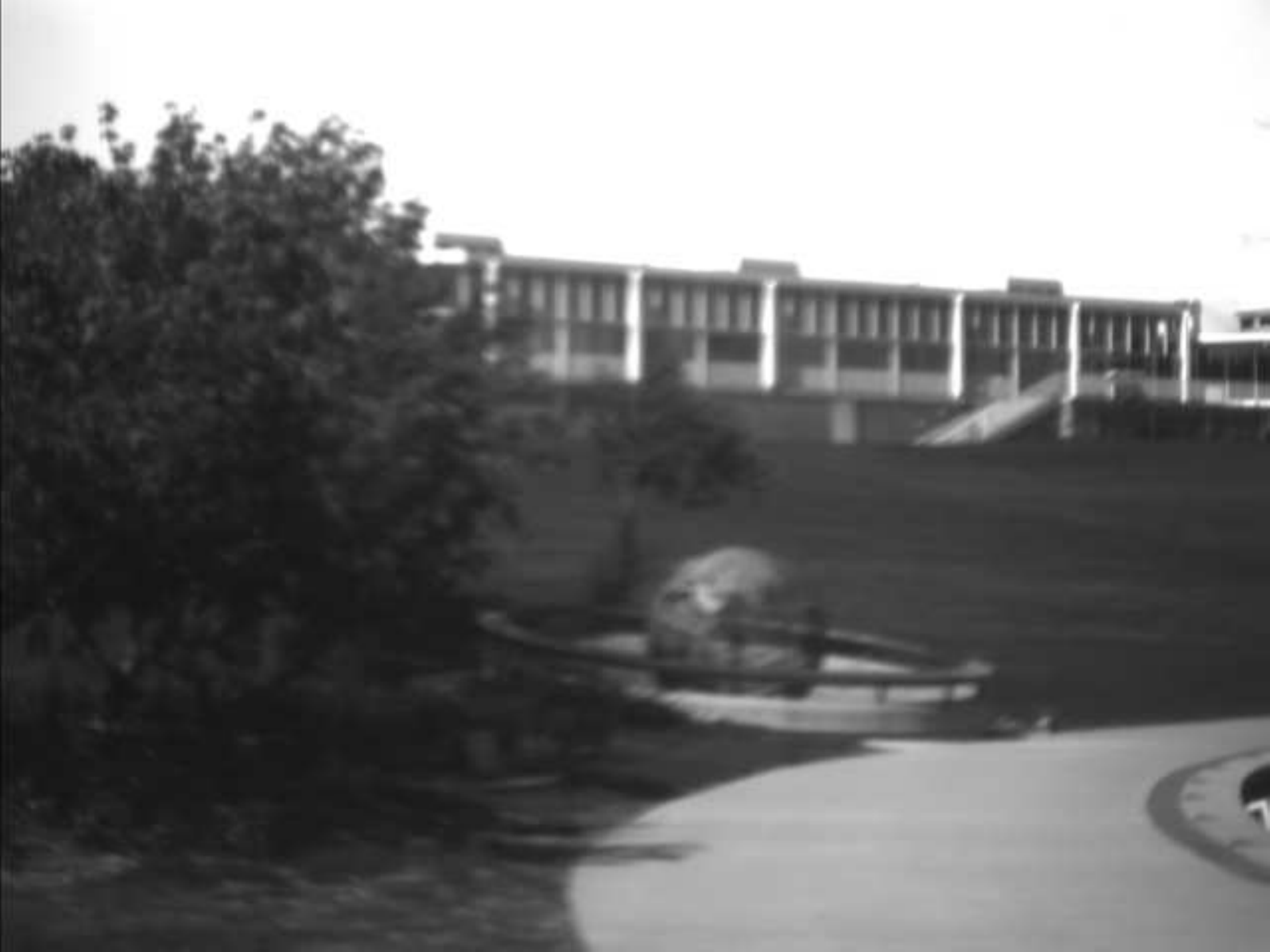}
\label{fig:D04band1}
}
\subfigure[]
{\includegraphics[width=0.3\linewidth]{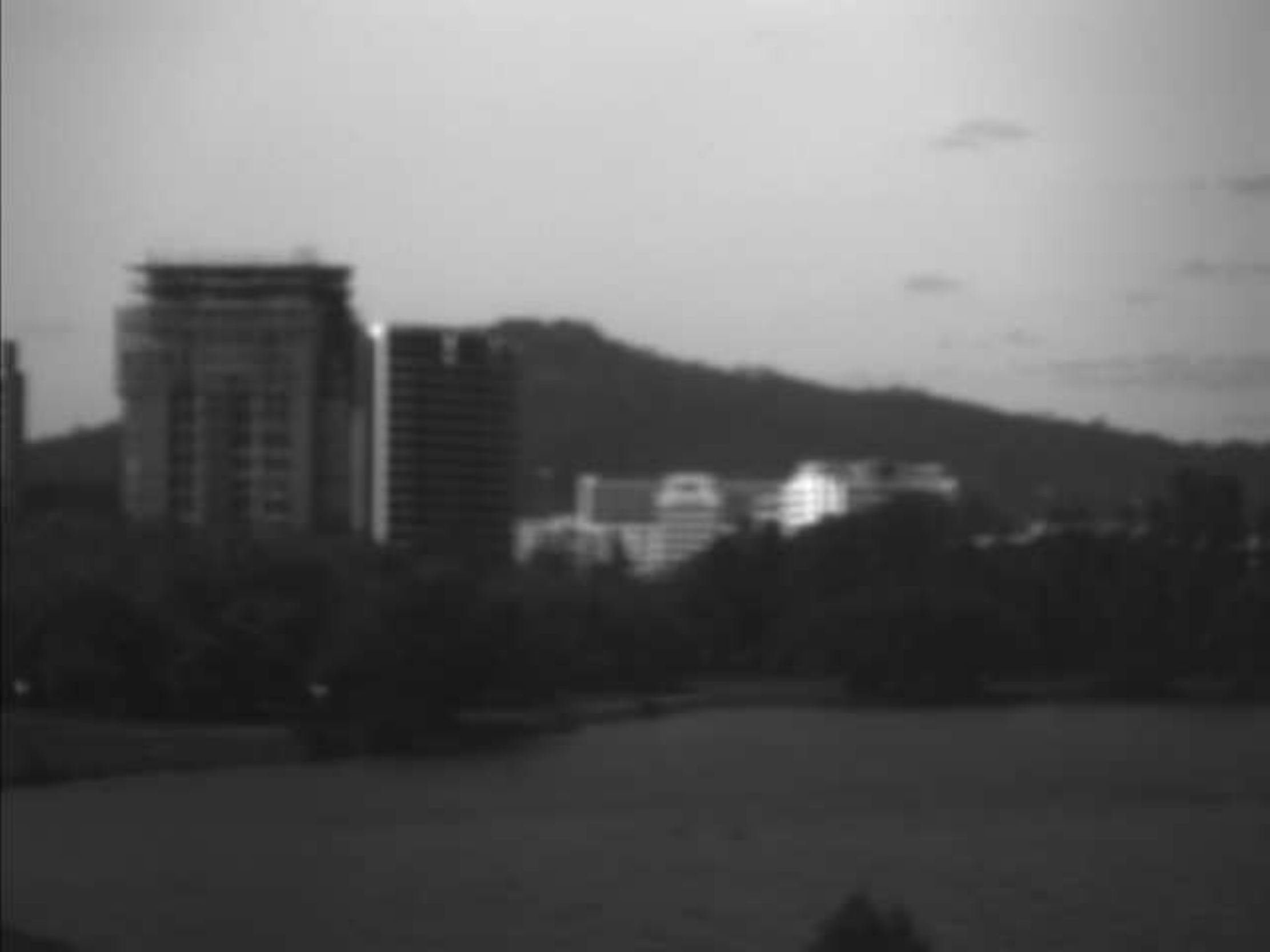}
\label{fig:G03band1}
}
\\
\subfigure[]{
\includegraphics[width=0.3\linewidth]{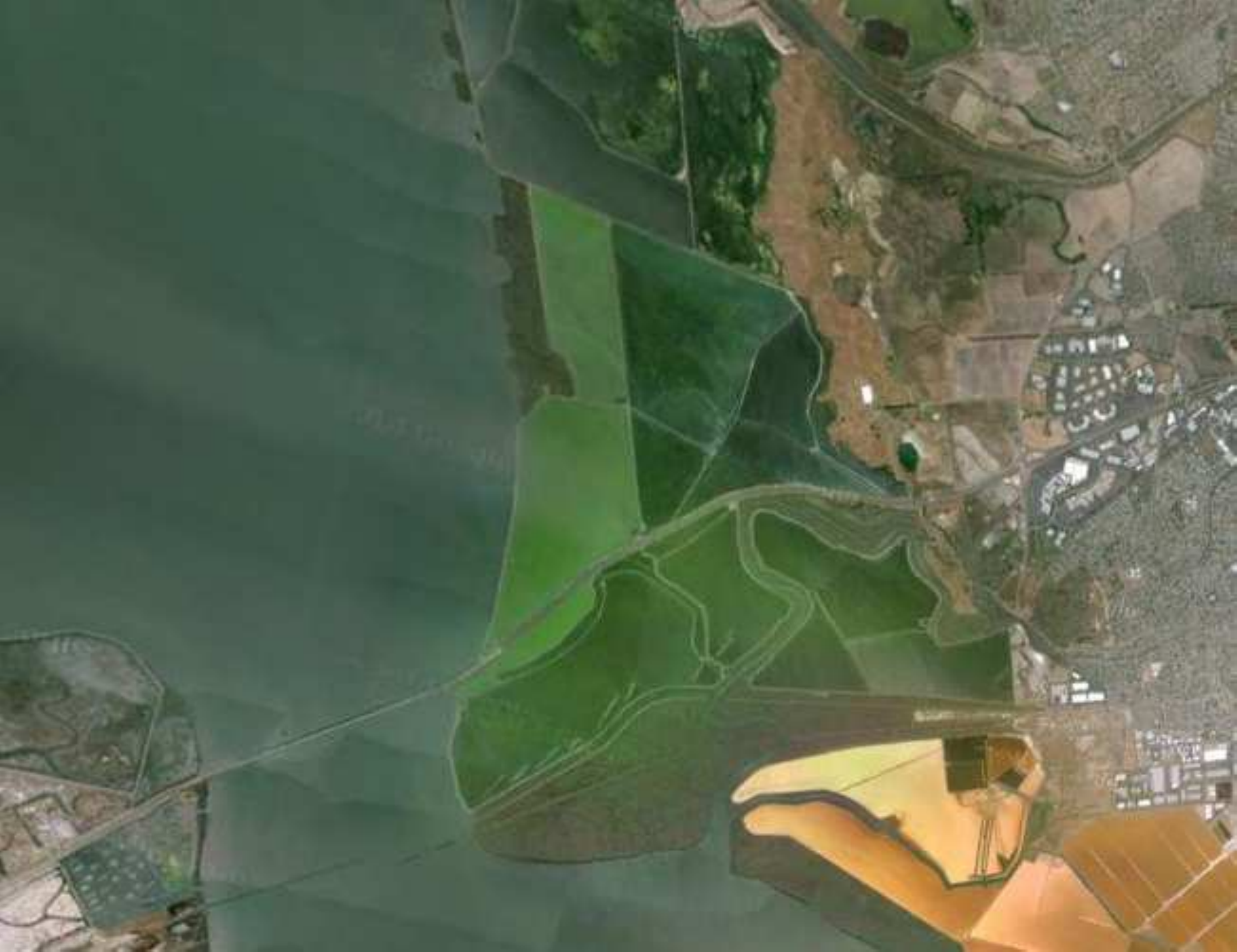}
\label{fig:RGB_moffett}
}
\subfigure[]{
\includegraphics[width=0.3\linewidth]{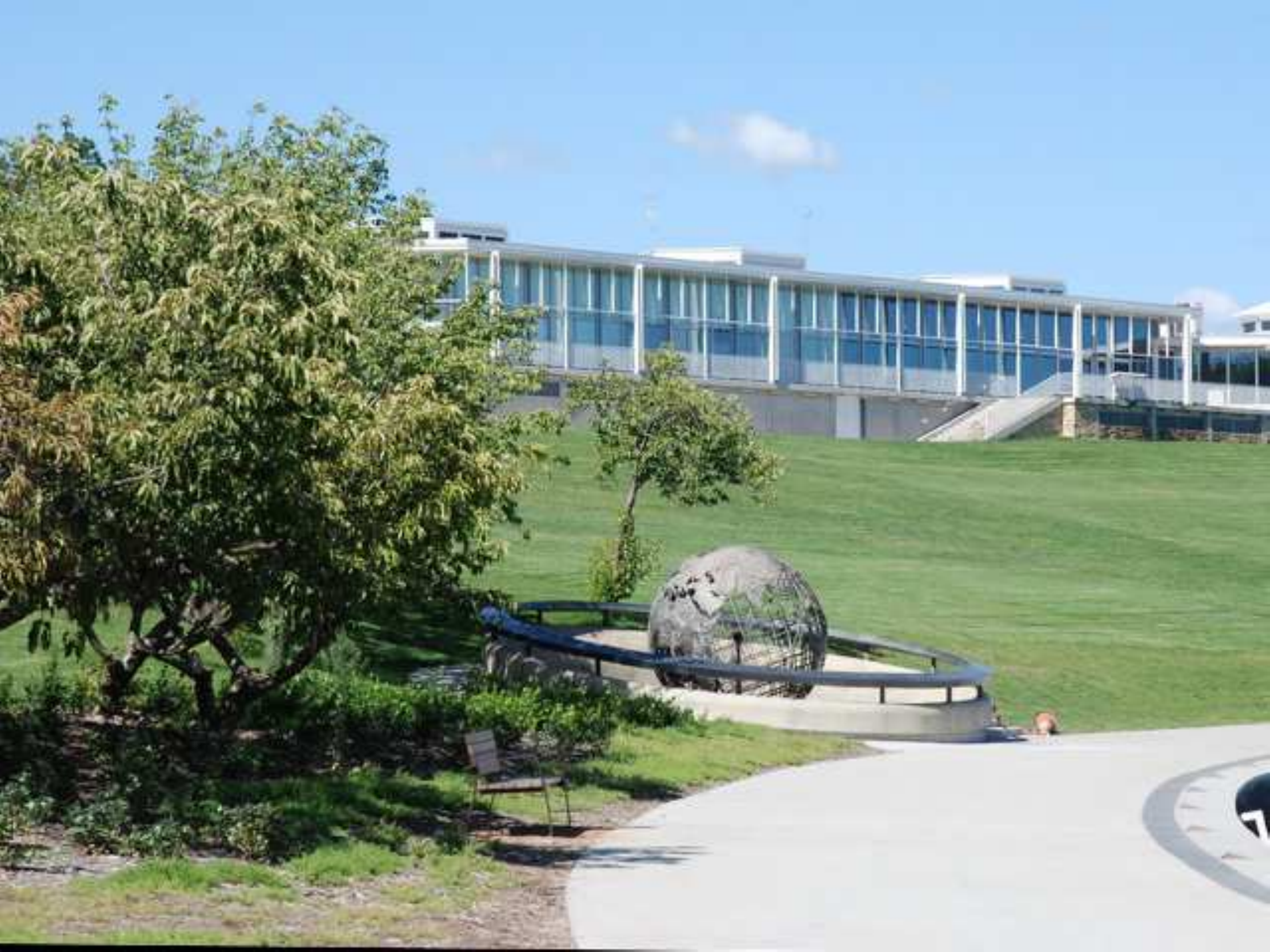}
\label{fig:RGB_D04}
}
\subfigure[]{
\includegraphics[width=0.29\linewidth]{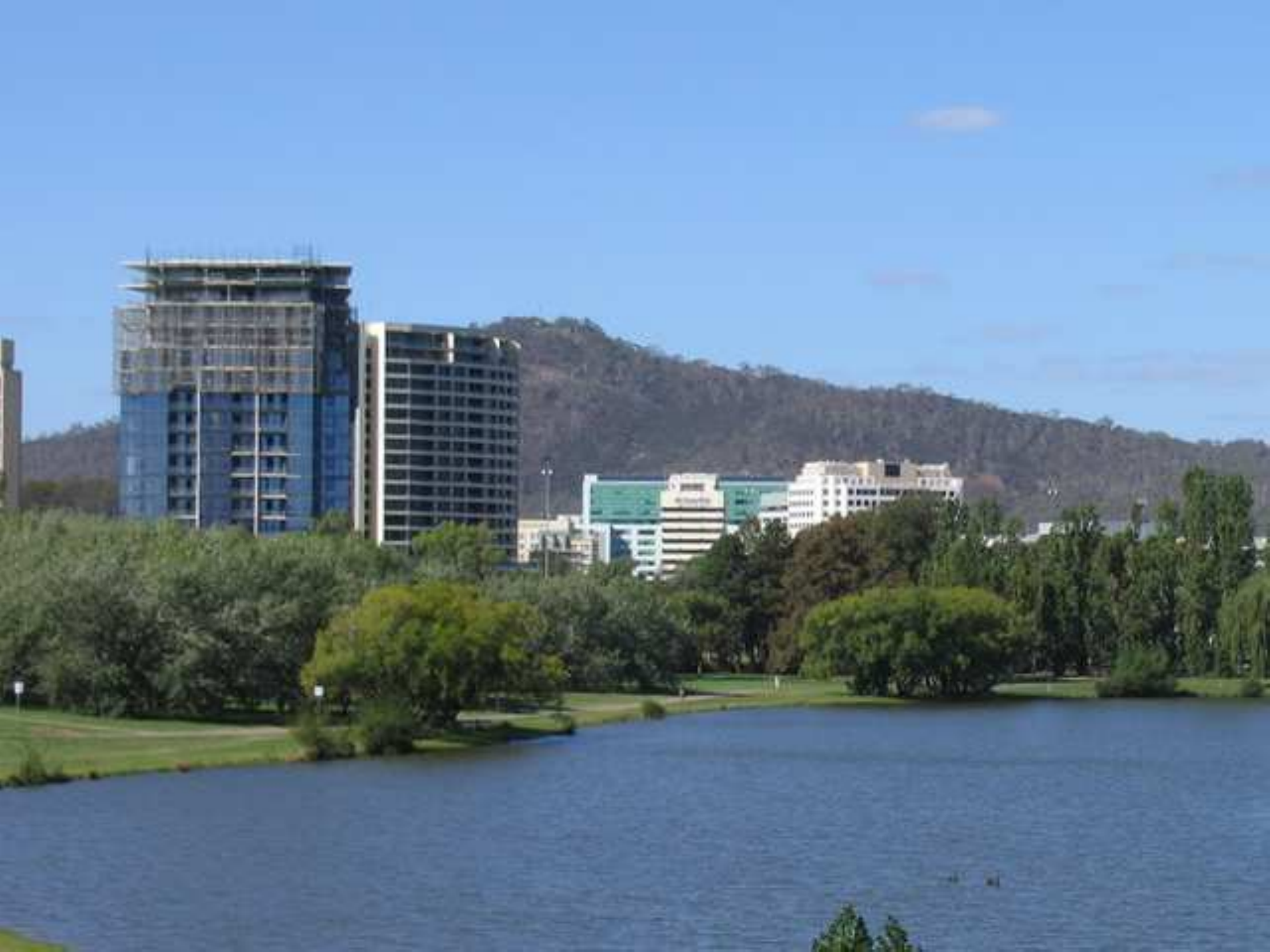}
\label{fig:RGB_G03}
}
\caption{Three HSIs and their corresponding RGB images. (a) The 50th band of the Moffett Field. (b) The first band of D04. (c) The first band of G03. (d) The RGB image of Moffett Field. (e) The RGB image of D03. (f) The RGB image of G03.}
\label{HSIandRGB}
\end{figure}
\begin{figure}[!tbp]
\centering
\subfigure[LP band selection]{
\includegraphics[angle=90,width=0.7\linewidth]{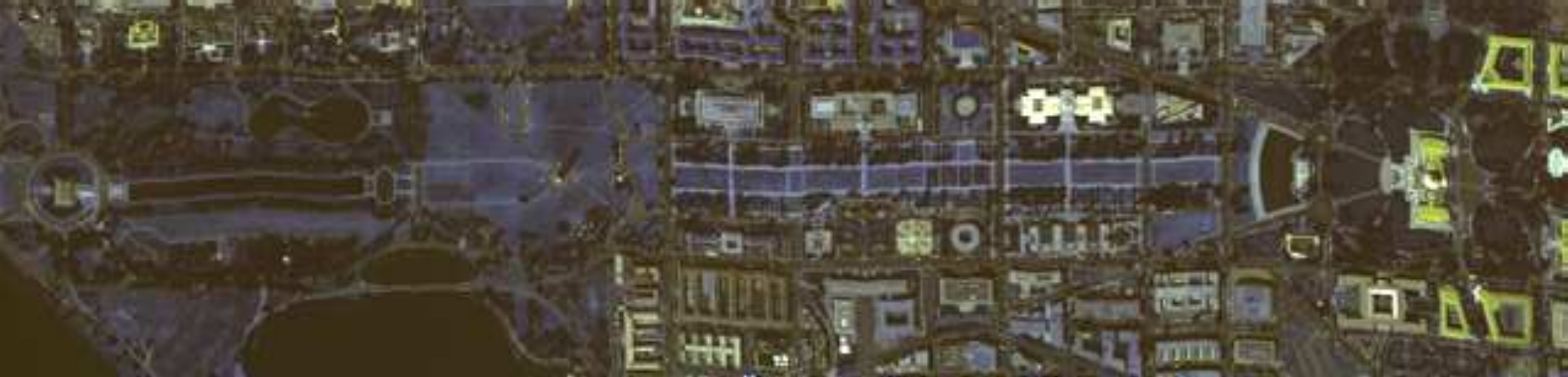}
\label{fig:DClpbandselection}
}
\subfigure[Manifold alignment]{
\includegraphics[angle=90,width=0.7\linewidth]{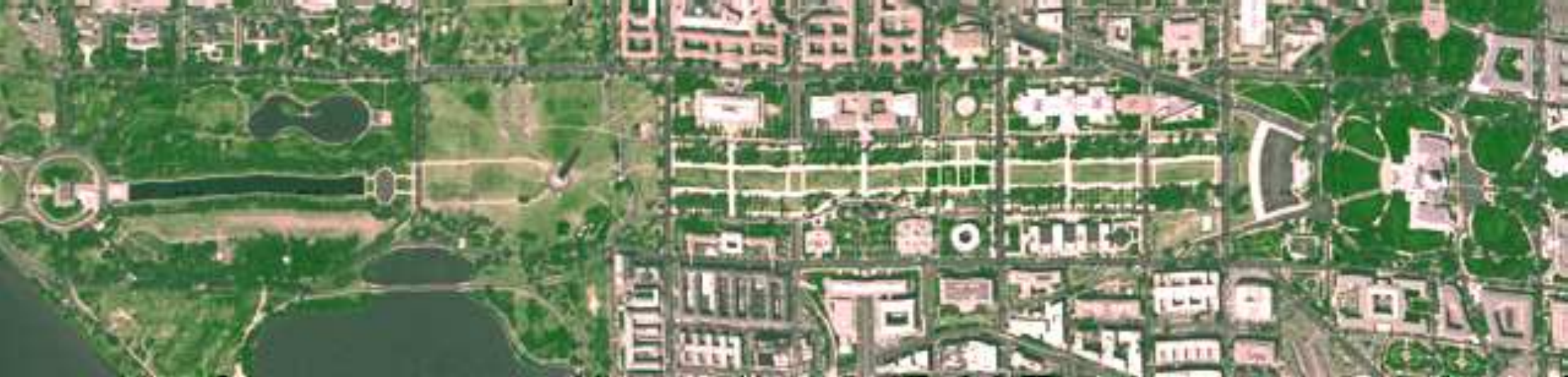}
\label{fig:DCmanifoldalignment}
}
\subfigure[Stretched CMF]{
\includegraphics[angle=90,width=0.7\linewidth]{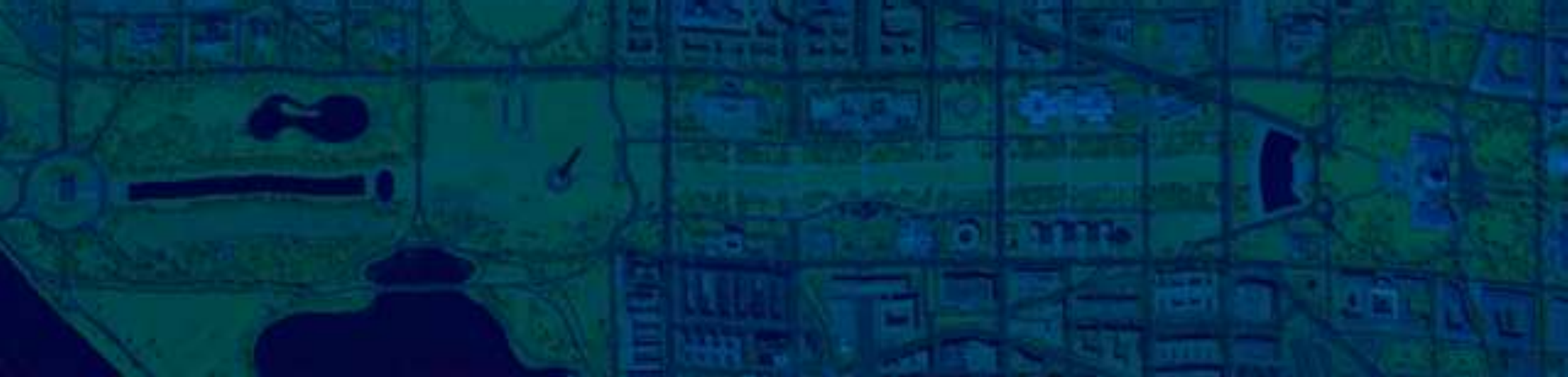}
\label{fig:DCCMF}
}
\subfigure[Bilateral filtering]{
\includegraphics[angle=90,width=0.7\linewidth]{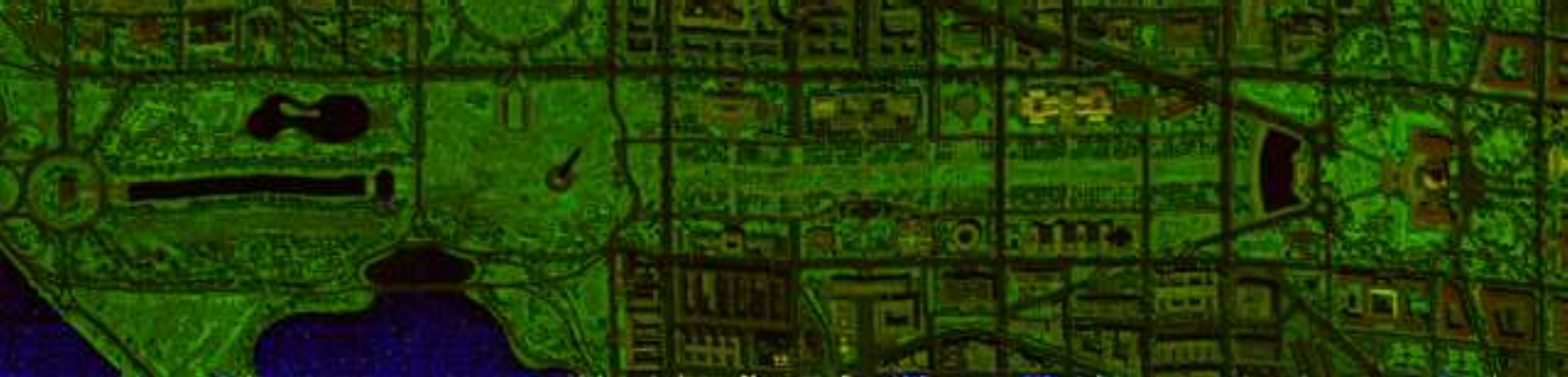}
\label{fig:DCBF}
}
\subfigure[Bicriteria optimization]{
\includegraphics[angle=90,width=0.7\linewidth]{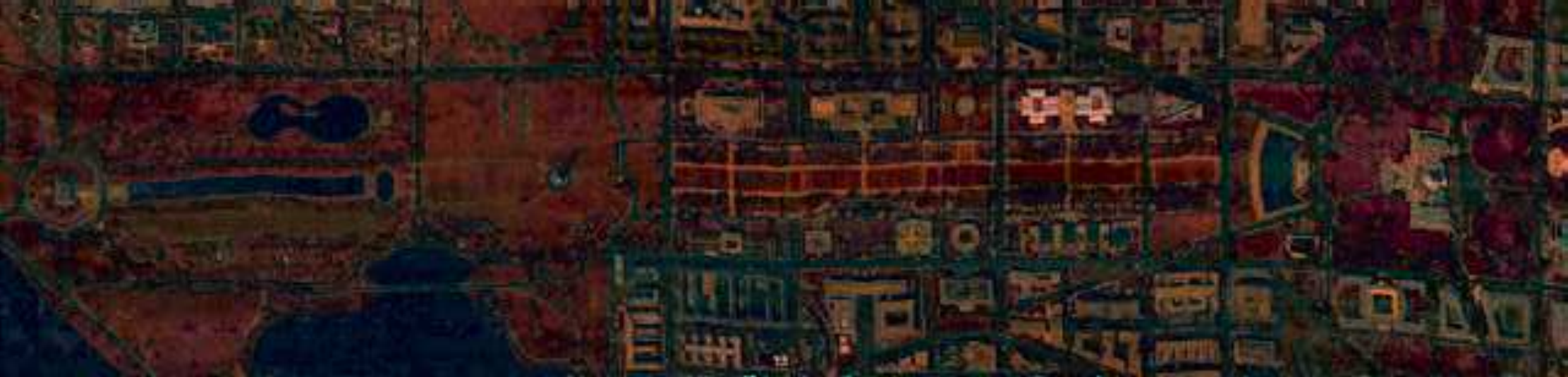}
\label{fig:DCBCOCDM}
}
\subfigure[MLS]{
\includegraphics[angle=90,width=0.7\linewidth]{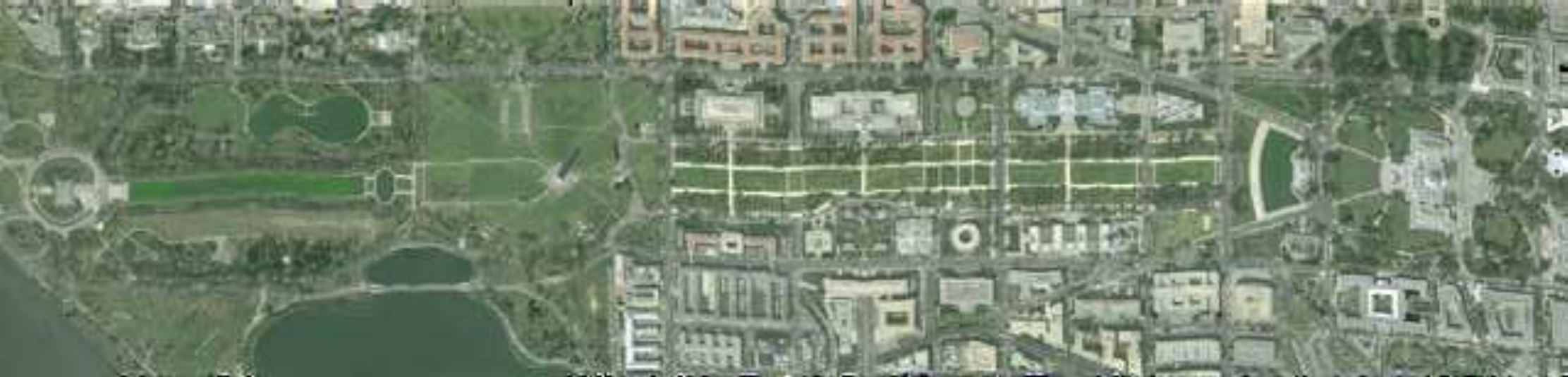}
\label{fig:DCMLS}
}
\caption{Visual comparison of different visualization approaches on the Washington DC Mall data set.}
\label{DCcomparison}
\end{figure}
Image registration is a common way to find the matching pixels between two images.
However, as an HSI and the corresponding RGB image are often captured by different cameras with different angles and image distortions, it is very difficult to obtain a precise image registration.
In this paper we propose a two-step coarse-to-fine searching process.
Firstly, the HSI and the RGB image are registered via a homography transform to get a coarse matching.
Scale-invariant feature transform (SIFT) is widely used to detect
the matching key points between images due to its robustness
to changes in scale, orientation, and illumination~\cite{lowe2004distinctive}.
In this paper, we first extract the SIFT key points from each band image of the HSI and the RGB image, and then select a few of the most similar pairs as the matching key points.
With the matching key points we estimate a homography transform using the Direct Linear Transformation (DLT) algorithm~\cite{hartley2003multiple}.

To obtain a more accurate matching, for each pixel in the HSI, we search for its final matching pixel in the neighborhood window of the current matching one with SIFT distance.
In our experiments, the size of the window is set to $9 \times 9$.
To find a set of matching pairs, we randomly select $1\%$ of pixels in the HSI and then find their matching pixels in the RGB image using the above process.
\subsection{Other Visualization Scenarios}
\emph{Reusing the Obtained Matching Pixels to Visualize Similar HSIs:}
Based on the fact that the same types of objects shall have similar
spectral responses with the same sensor,
the matching pixels between a pair of HSI and RGB image can be directly applied to visualize other semantically similar HSIs captured by the same imaging sensor.
This scenario is very helpful for a quick overview of a batch of similar HSIs since the matching process between the HSI and the RGB image only needs to be done once.

\emph{Displaying an HSI Aided by an RGB Image Captured on a Different Site:}
The MLS-based visualization method only requires a small number of matching pixel pairs for color constraint.
Thus in the absence of a corresponding RGB image captured on the same site as the input HSI, a semantically similar RGB image can be used to help visualization.
In this case, interactive tools can be developed for users to manually select a number of matching pixels between the two images.

\begin{figure}[t]
\centering
\subfigure[LP band selection.]{
\includegraphics[width=0.3\linewidth]{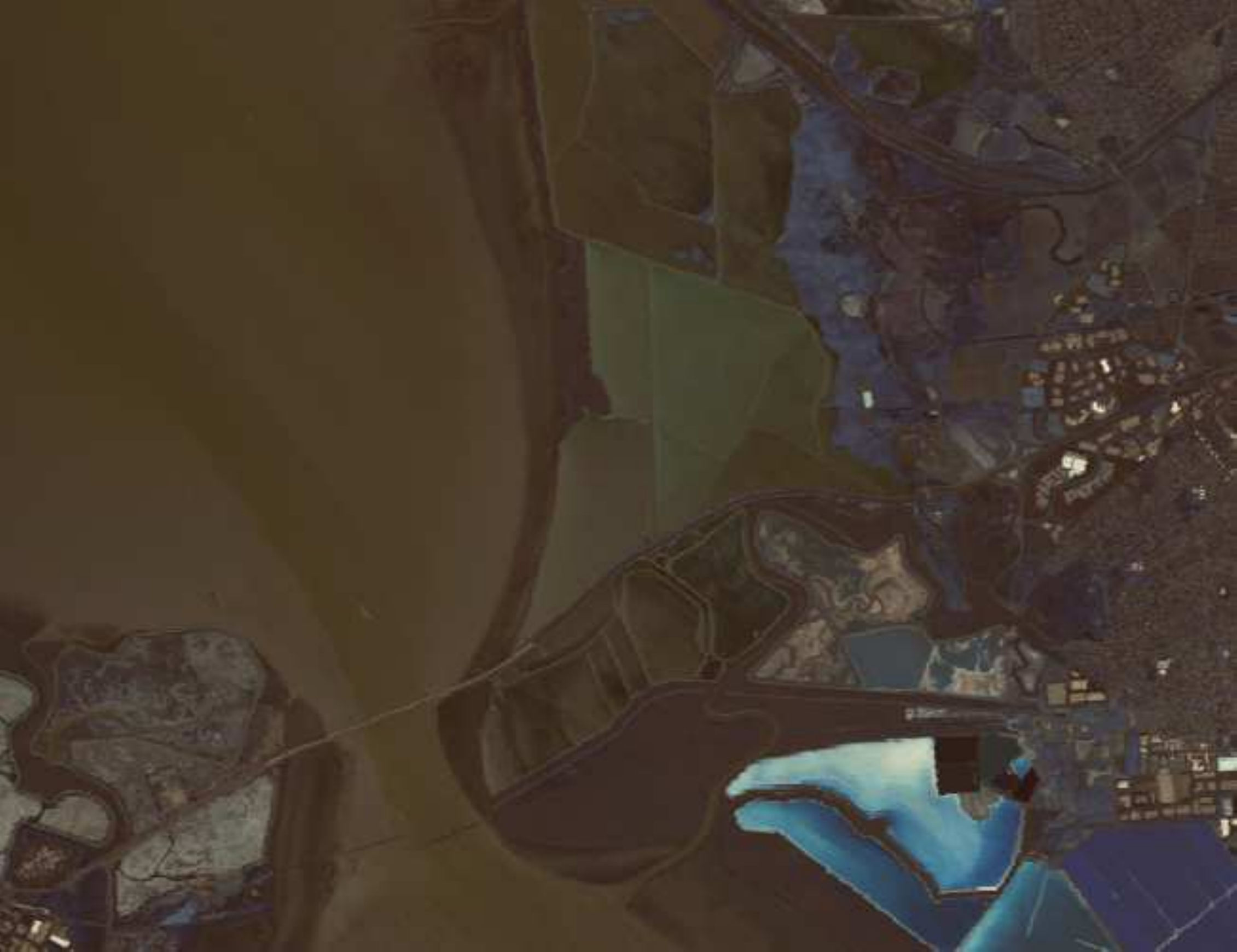}
}
\subfigure[Manifold alignment]{
\includegraphics[width=0.3\linewidth]{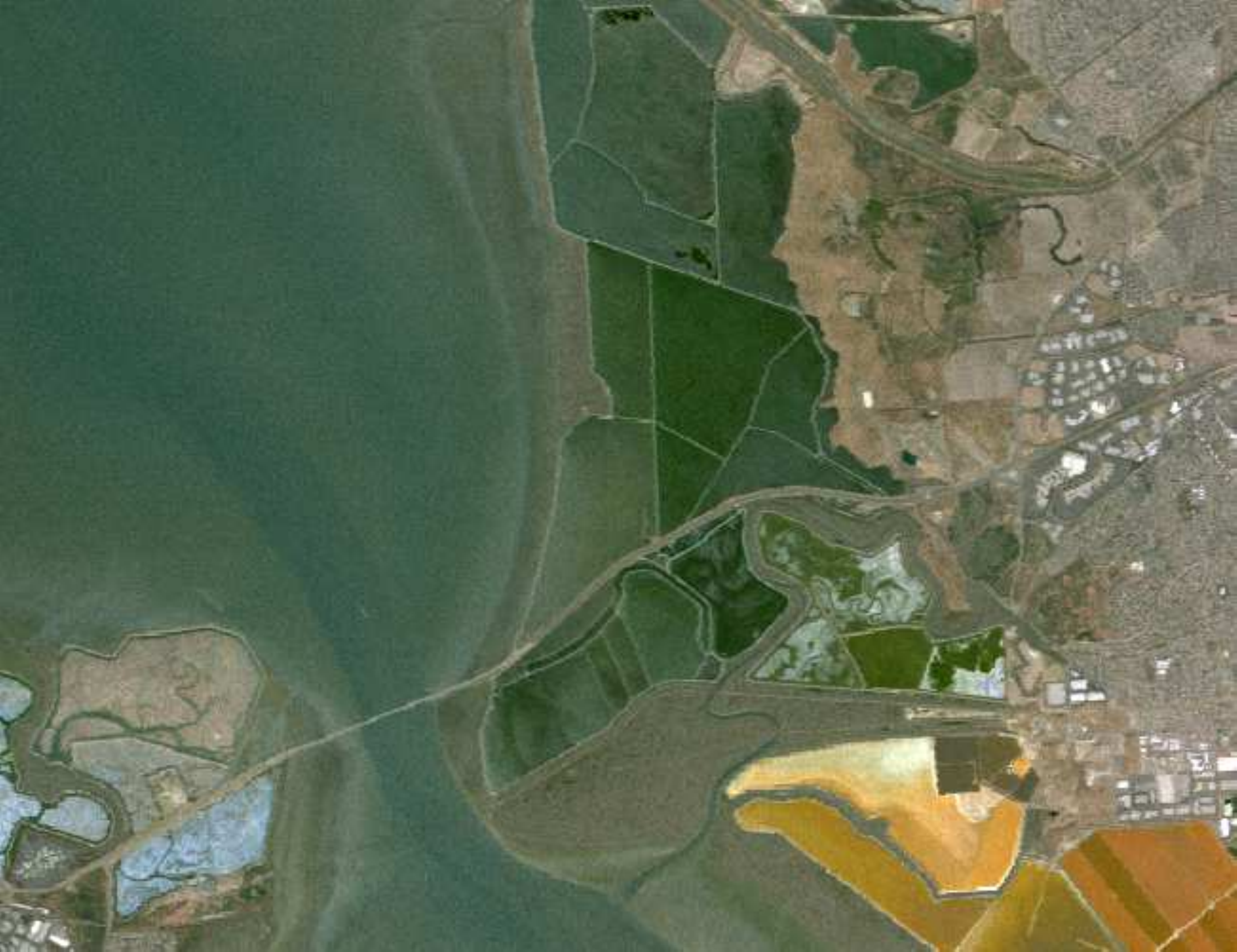}
}
\subfigure[Stretched CMF]{
\includegraphics[width=0.3\linewidth]{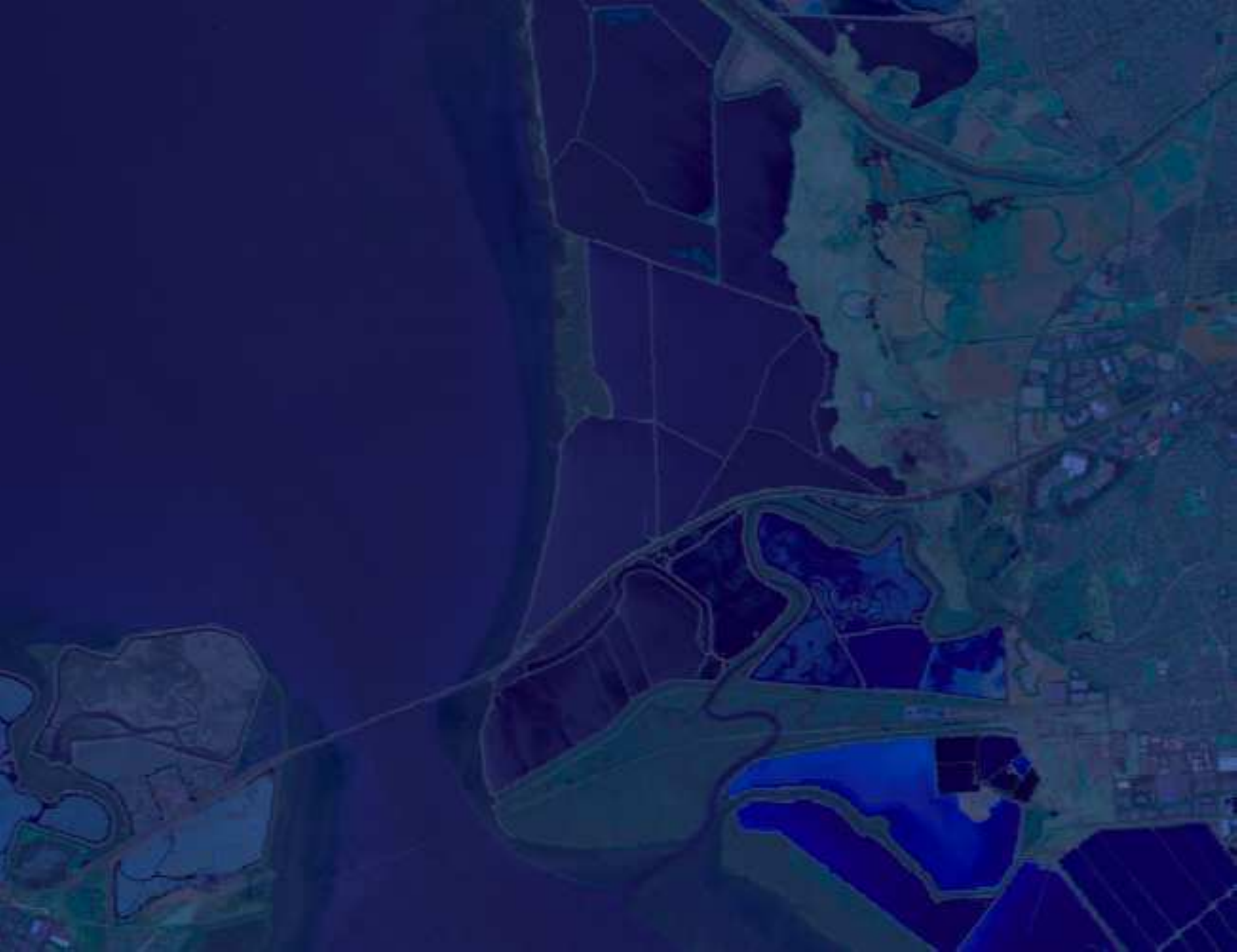}
}
\subfigure[Bilateral filtering]{
\includegraphics[width=0.3\linewidth]{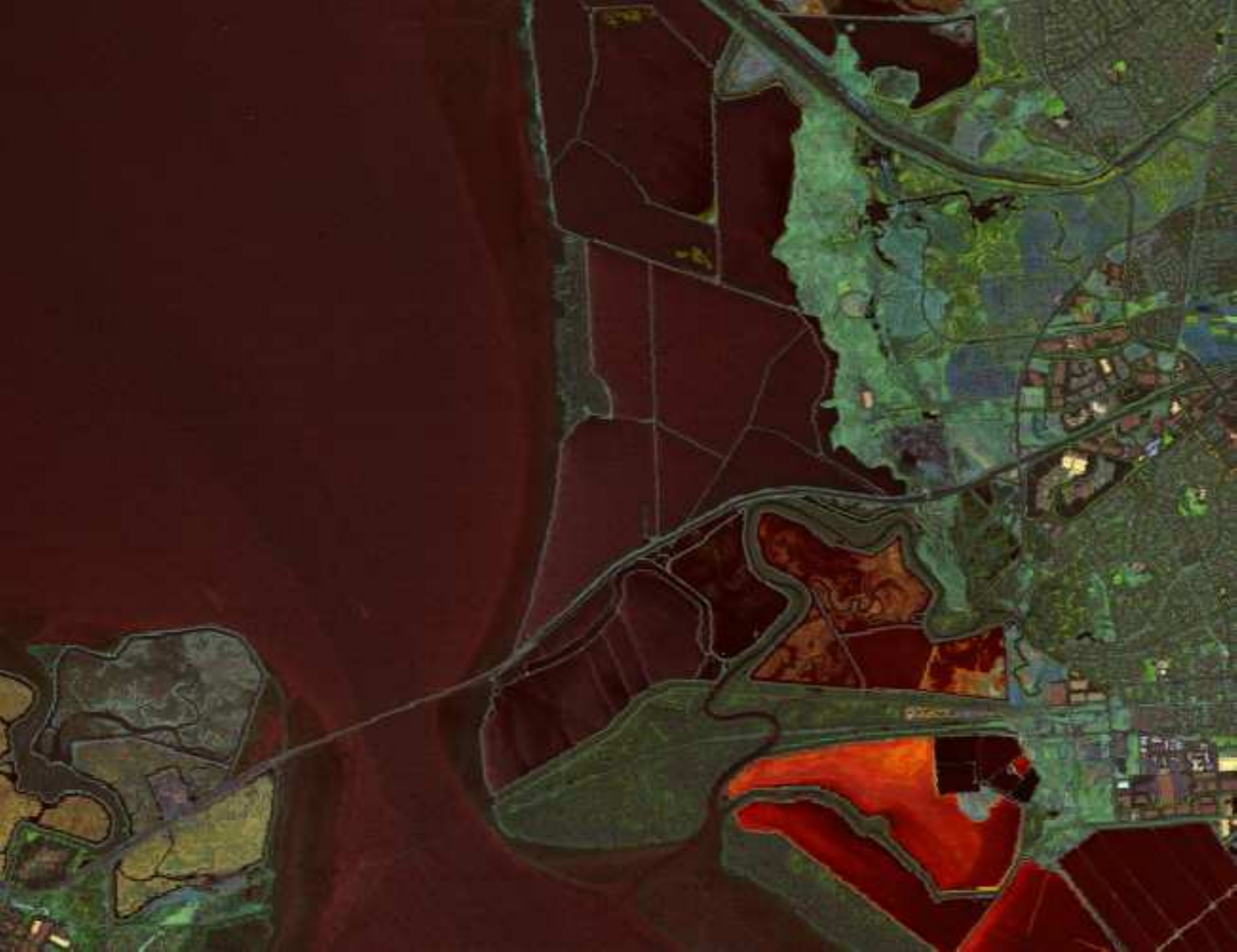}
}
\subfigure[Bicriteria optimization]{
\includegraphics[width=0.3\linewidth]{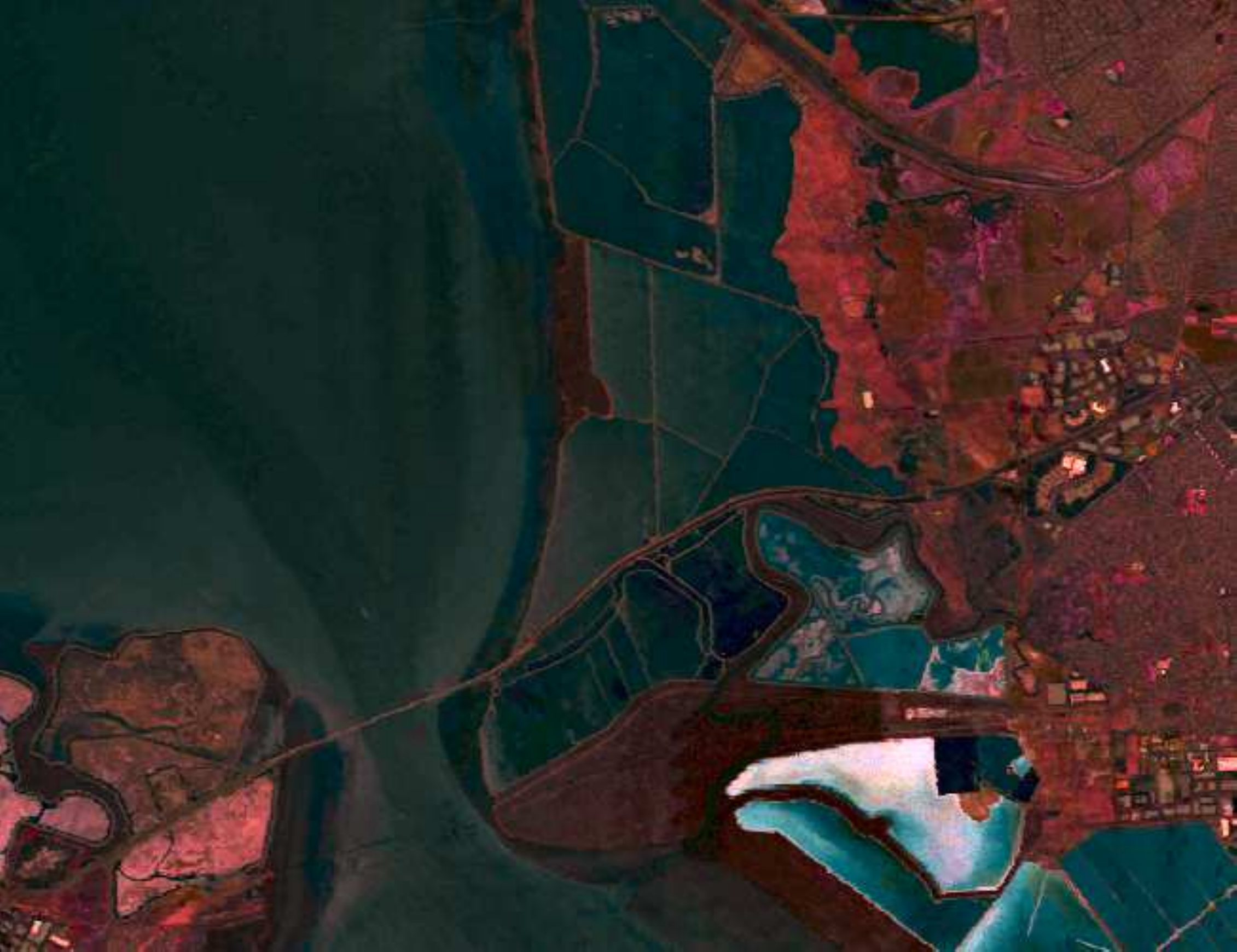}
}
\subfigure[MLS]{
\includegraphics[width=0.3\linewidth]{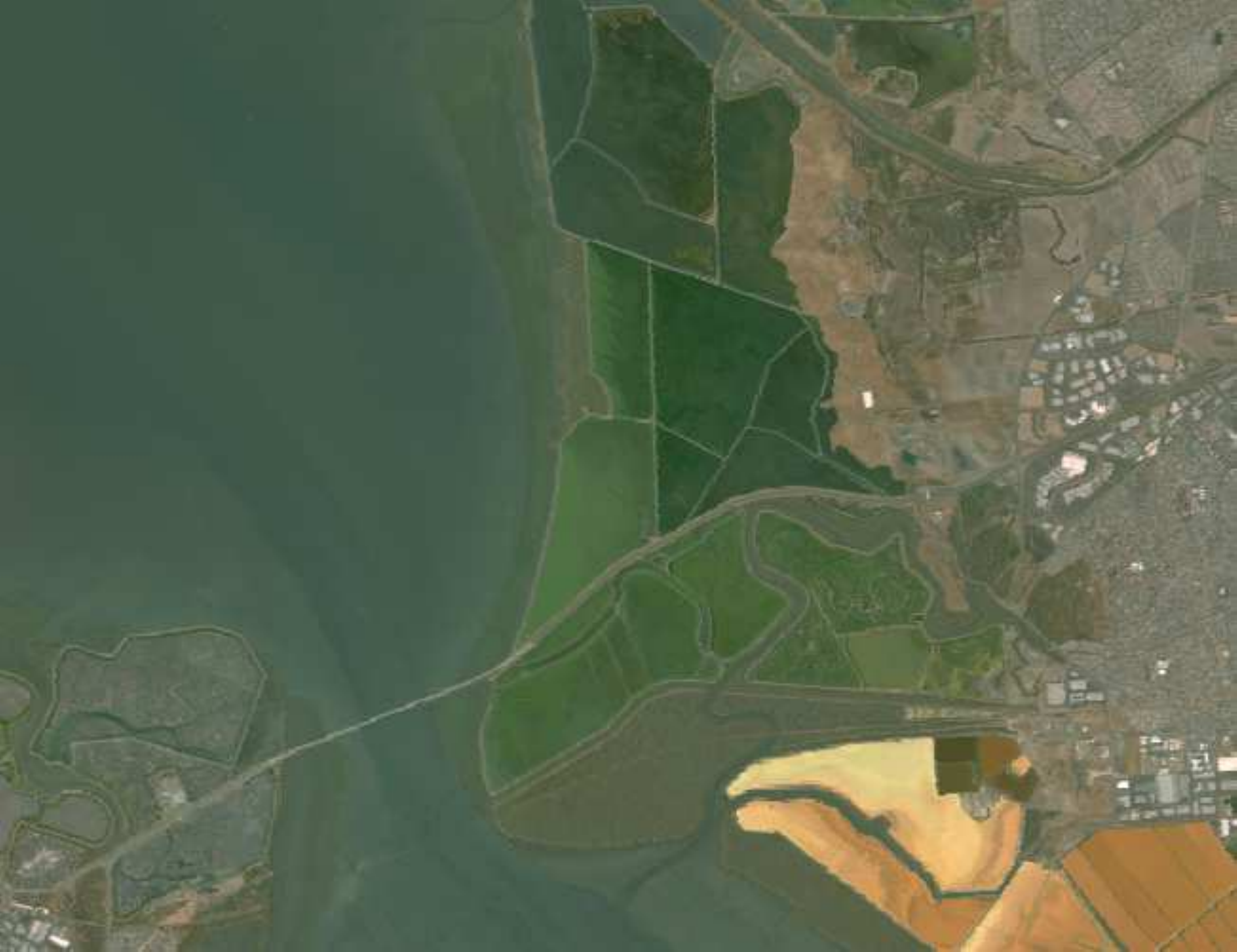}
}
\caption{Visual comparison of different visualization approaches on the Moffett Field data set.}\label{MoffettExperiments}
\end{figure}

\begin{figure}[t]
\centering
\subfigure[LP band selection]{
\includegraphics[width=0.3\linewidth]{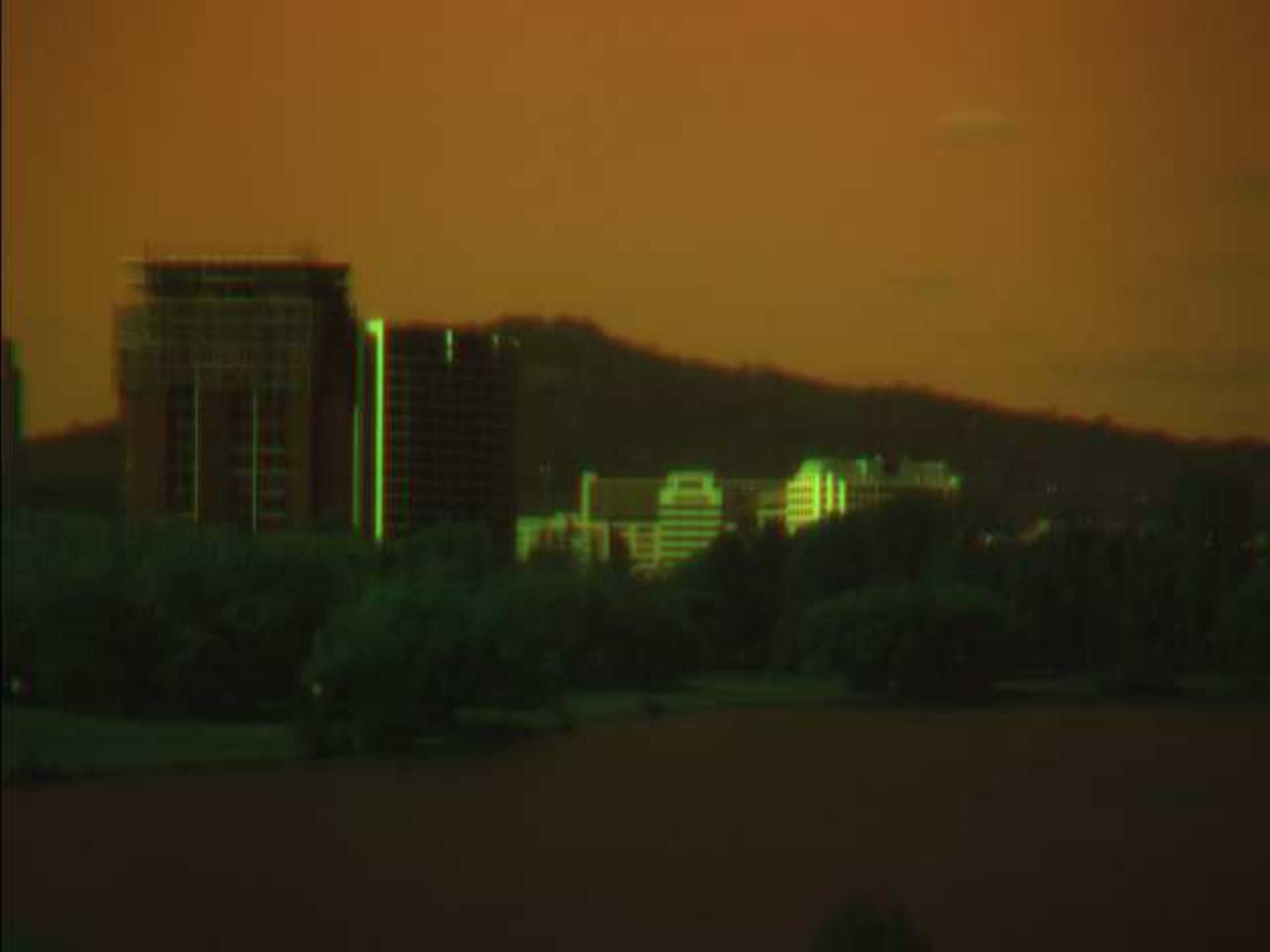}
}
\subfigure[Manifold alignment]{
\includegraphics[width=0.3\linewidth]{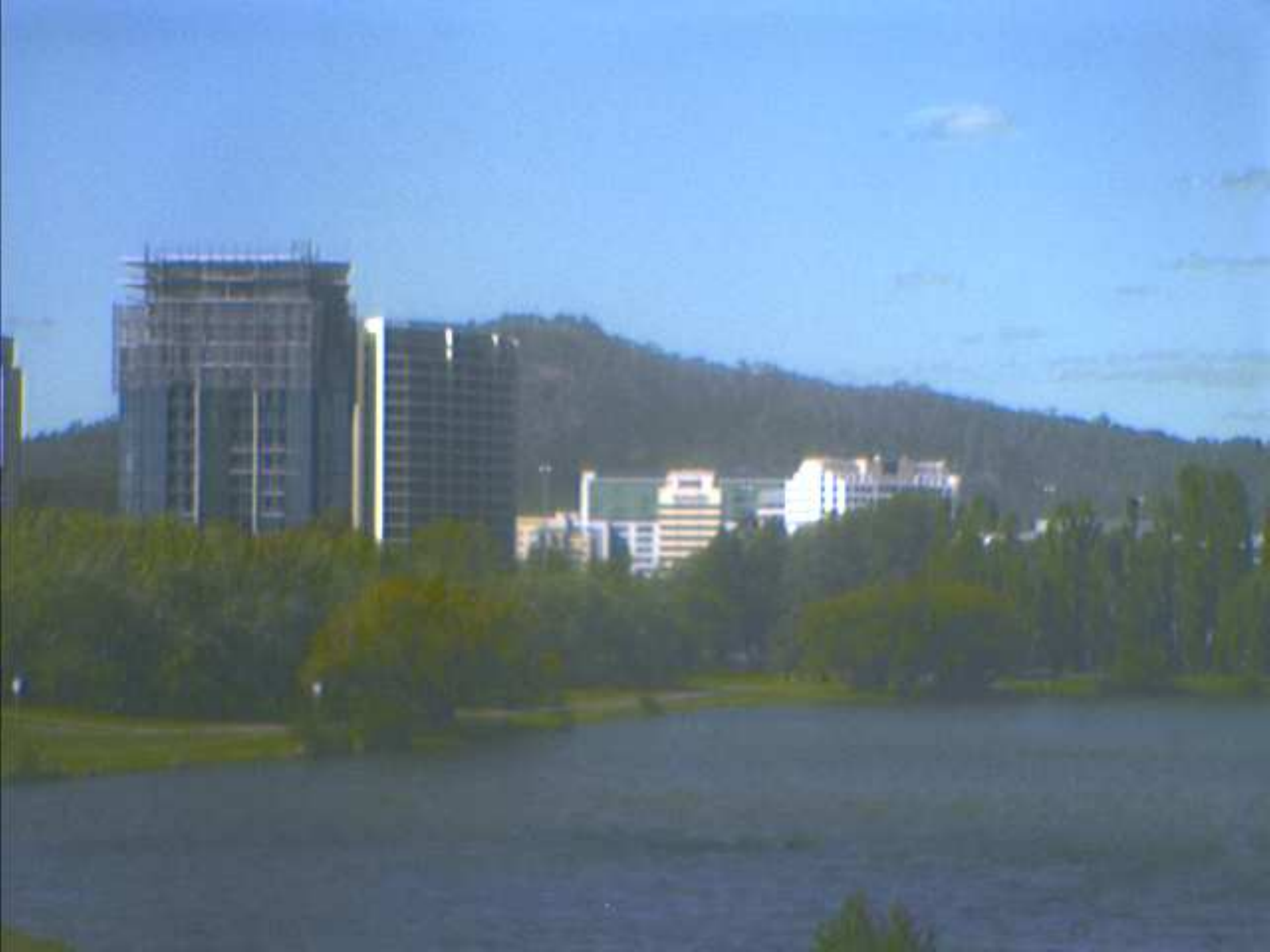}
}
\subfigure[Stretched CMF]{
\includegraphics[width=0.3\linewidth]{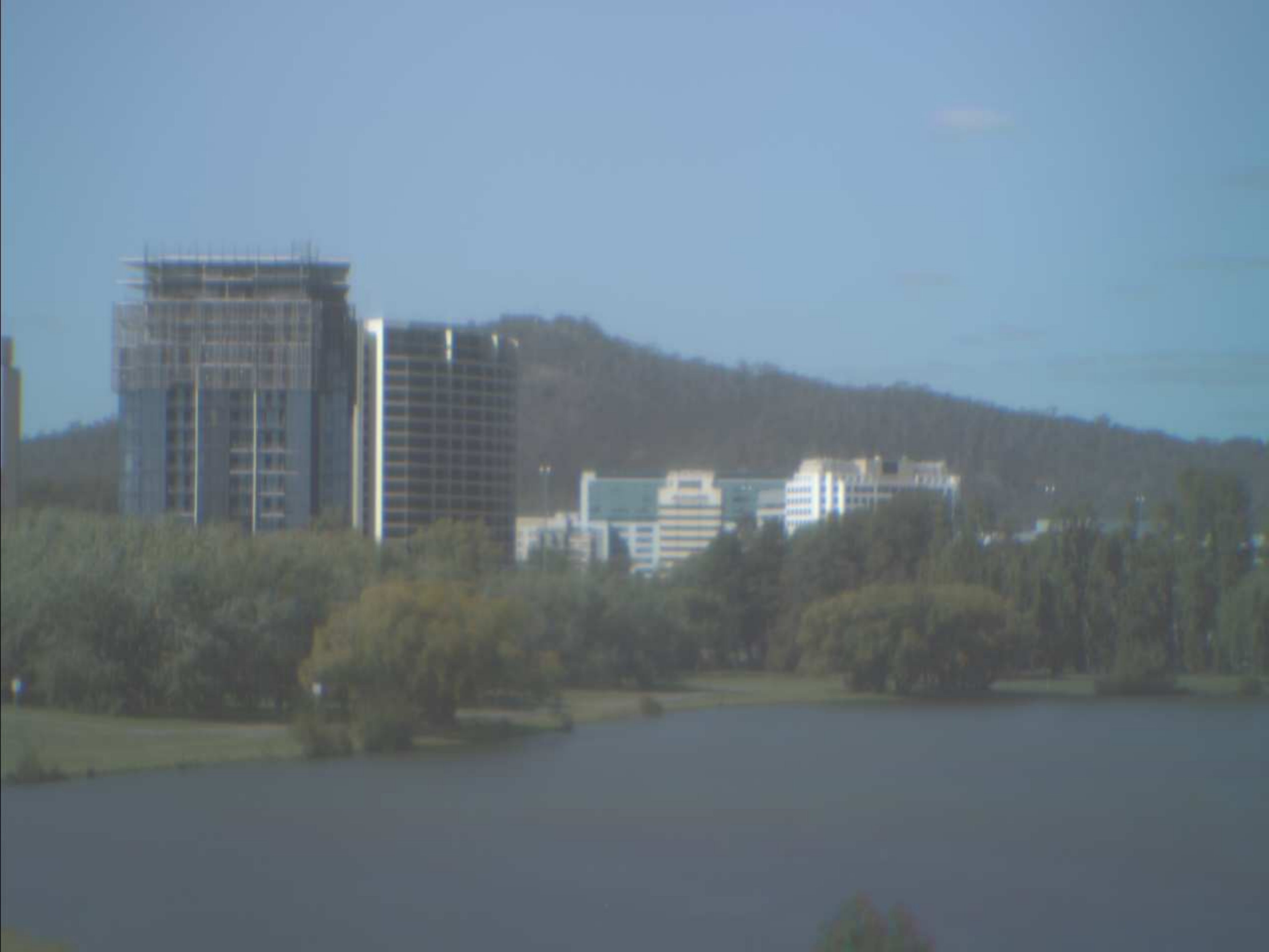}
}
\subfigure[Bilateral filtering]{
\includegraphics[width=0.3\linewidth]{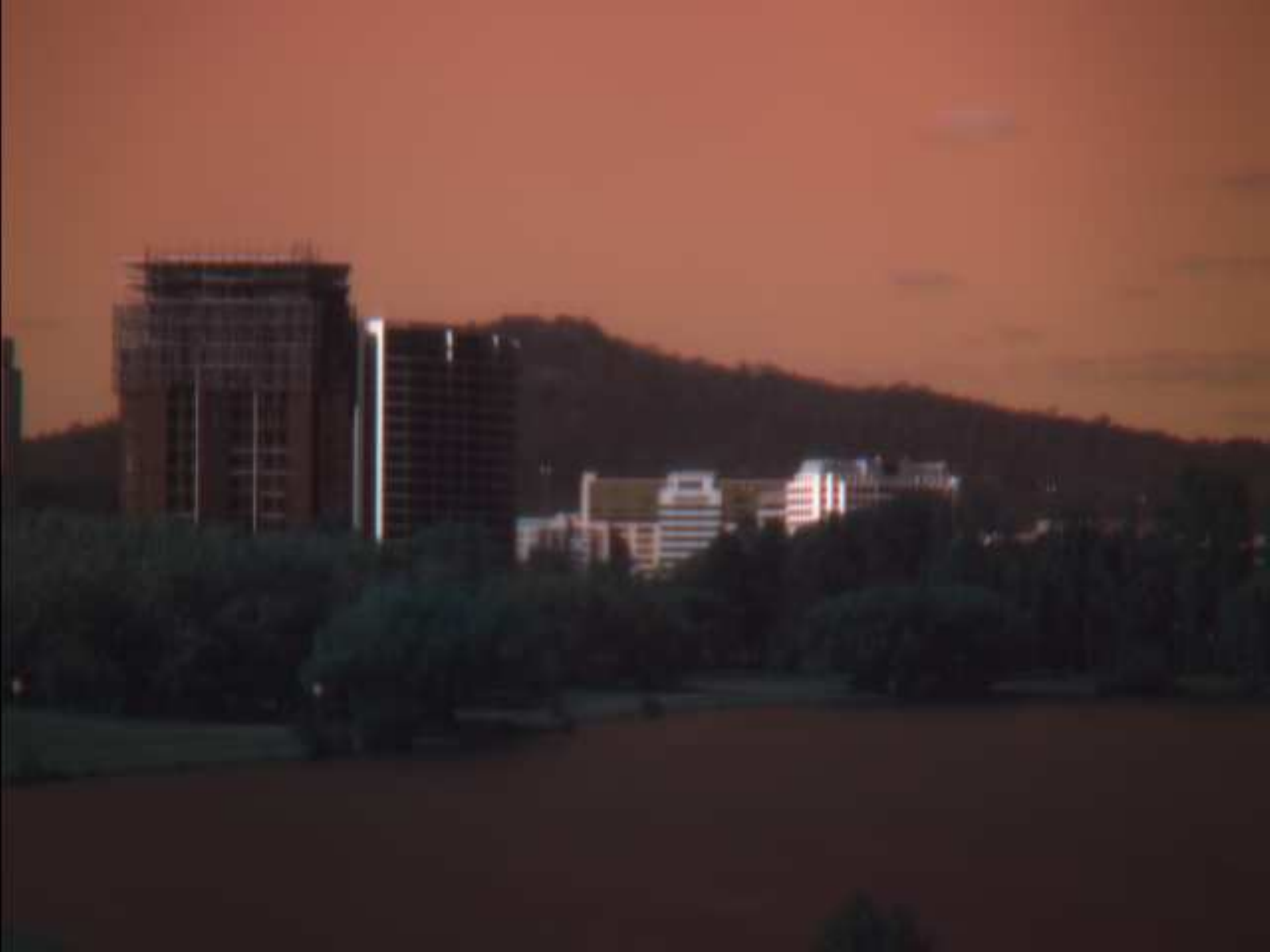}
}
\subfigure[Bicriteria optimization]{
\includegraphics[width=0.3\linewidth]{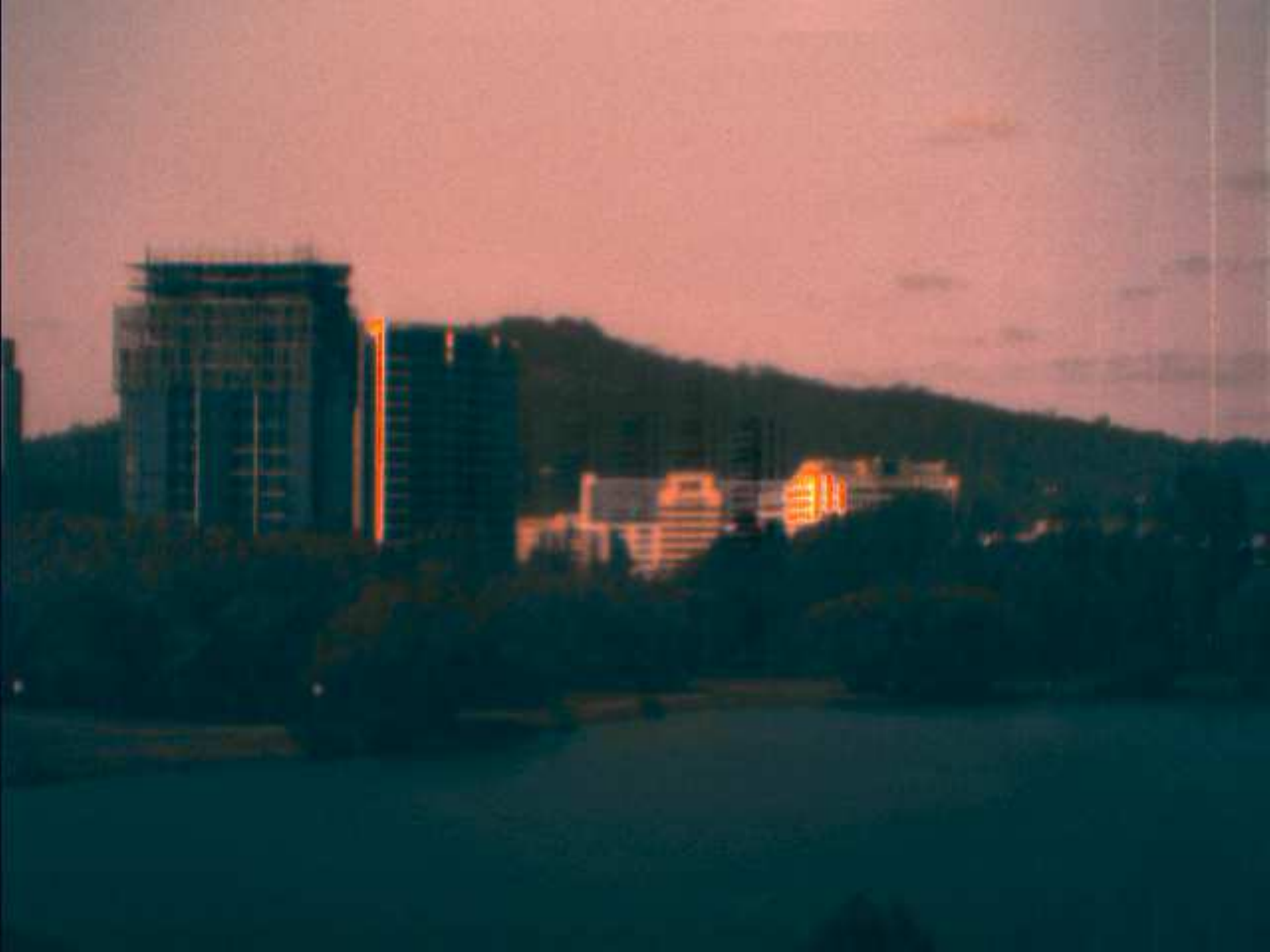}
}
\subfigure[MLS]{
\includegraphics[width=0.3\linewidth]{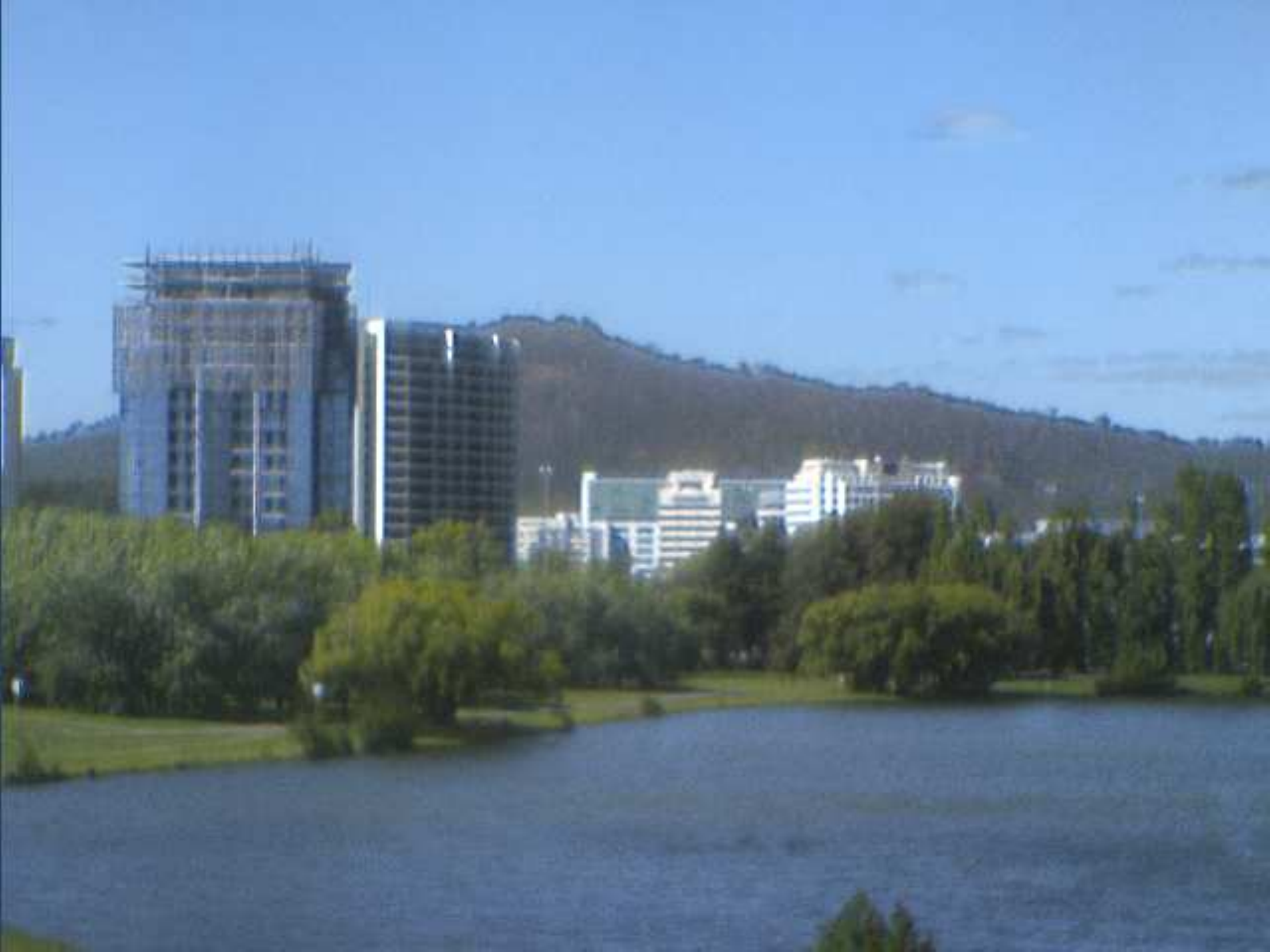}
}
\caption{Visual comparison of different visualization approaches on the G03 data set.}
\label{G03comparison}
\end{figure}
\begin{figure}[t]
\centering
\subfigure[LP band selection]{
\includegraphics[width=0.3\linewidth]{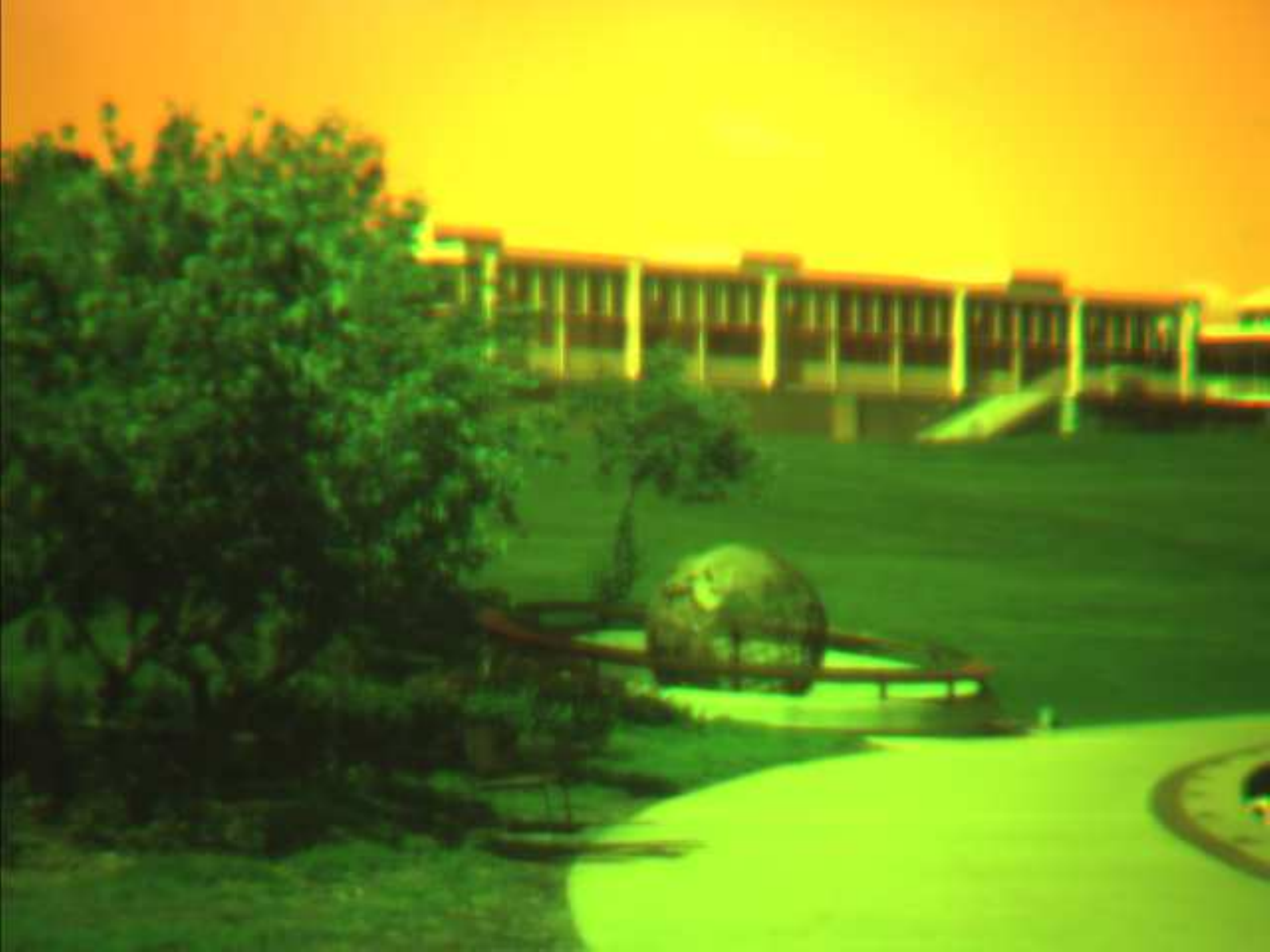}
\label{d04bandselection}
}
\subfigure[Manifold alignment]{
\includegraphics[width=0.3\linewidth]{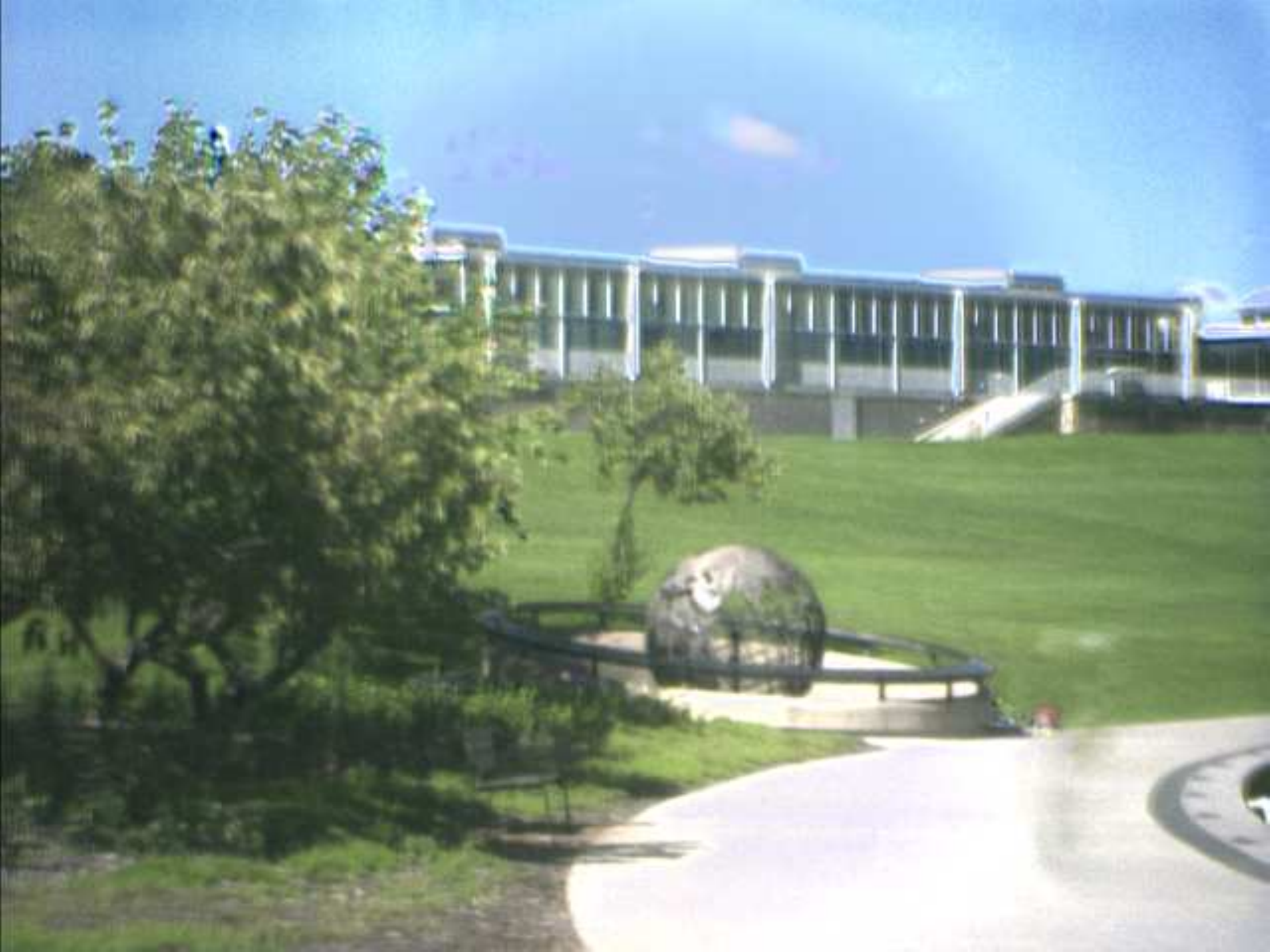}
\label{d04manifoldalignment}
}
\subfigure[Stretched CMF]{
\includegraphics[width=0.3\linewidth]{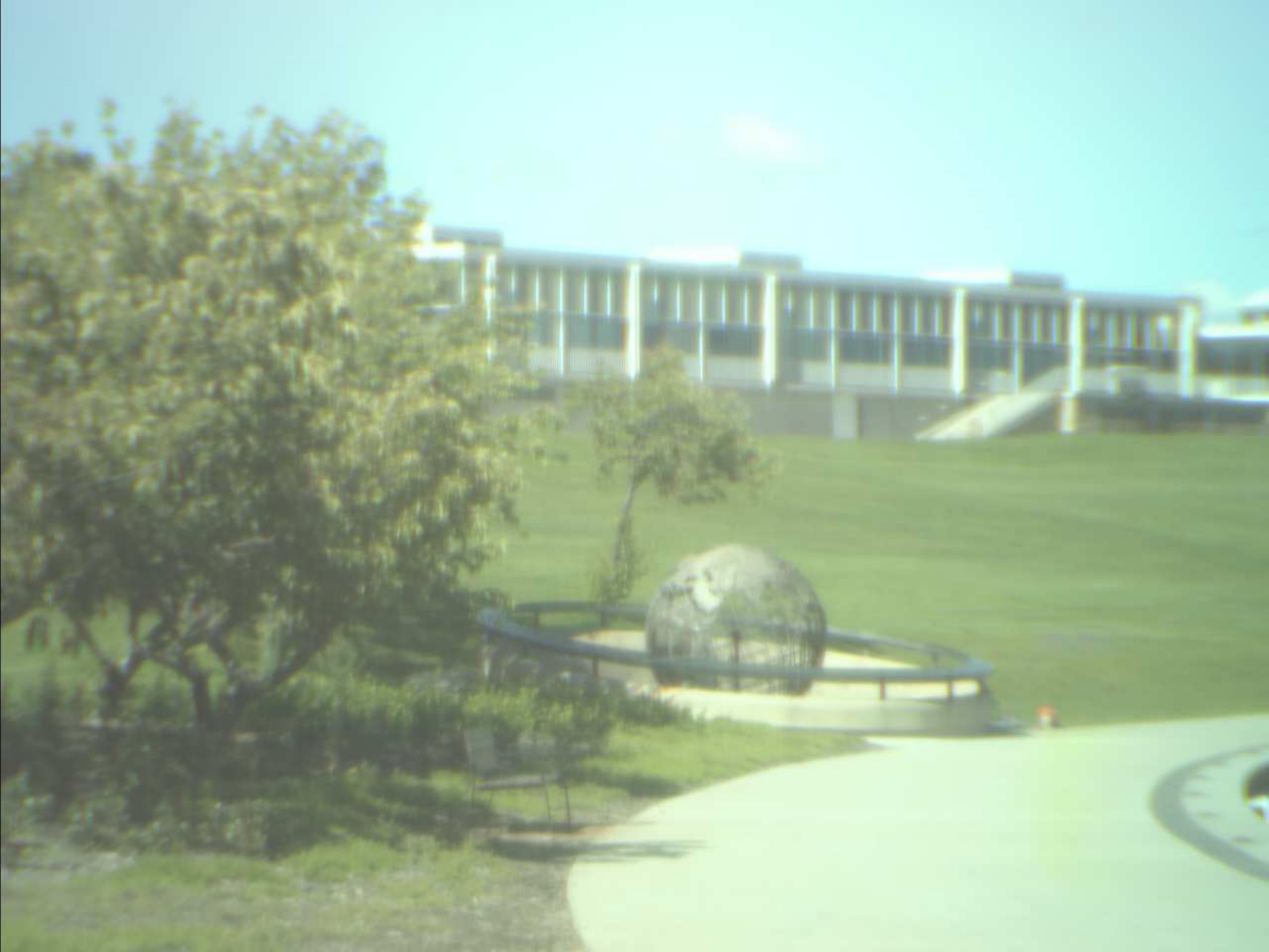}
\label{d04CMF}
}
\subfigure[Bilateral filtering]{
\includegraphics[width=0.3\linewidth]{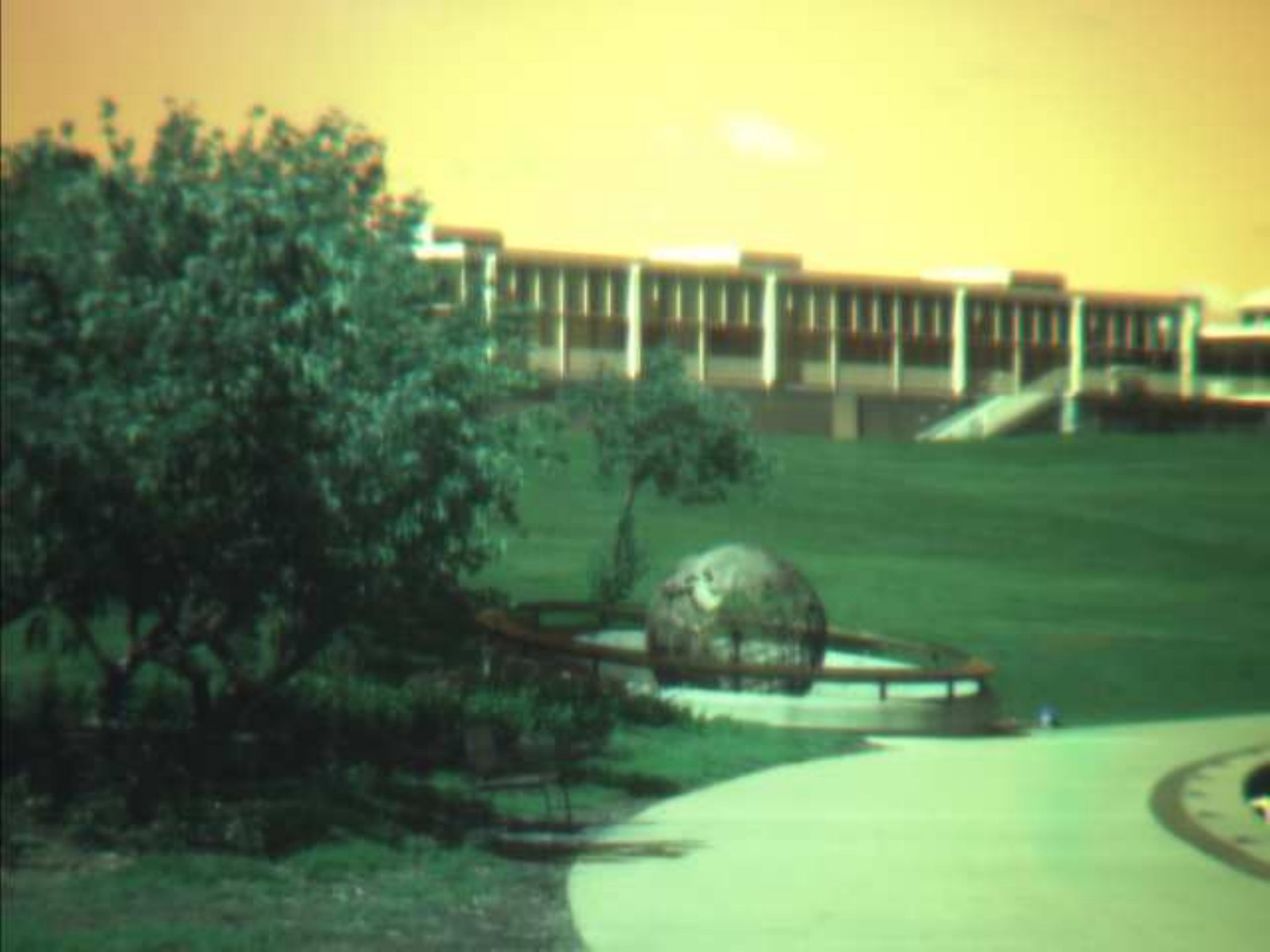}
\label{d04BF}
}
\subfigure[Bicriteria optimization]{
\includegraphics[width=0.3\linewidth]{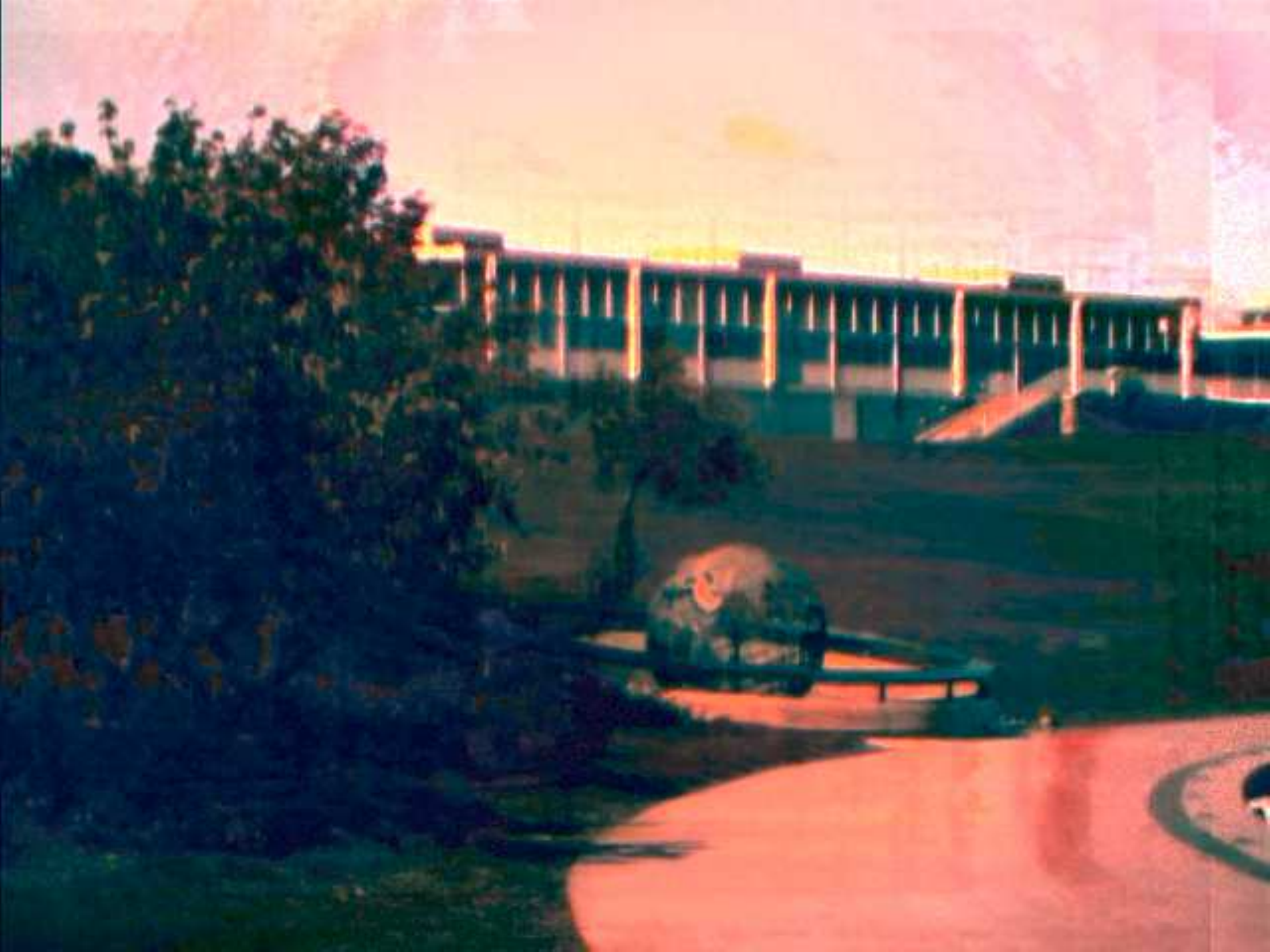}
\label{d04DCOCDM}
}
\subfigure[MLS]{
\includegraphics[width=0.3\linewidth]{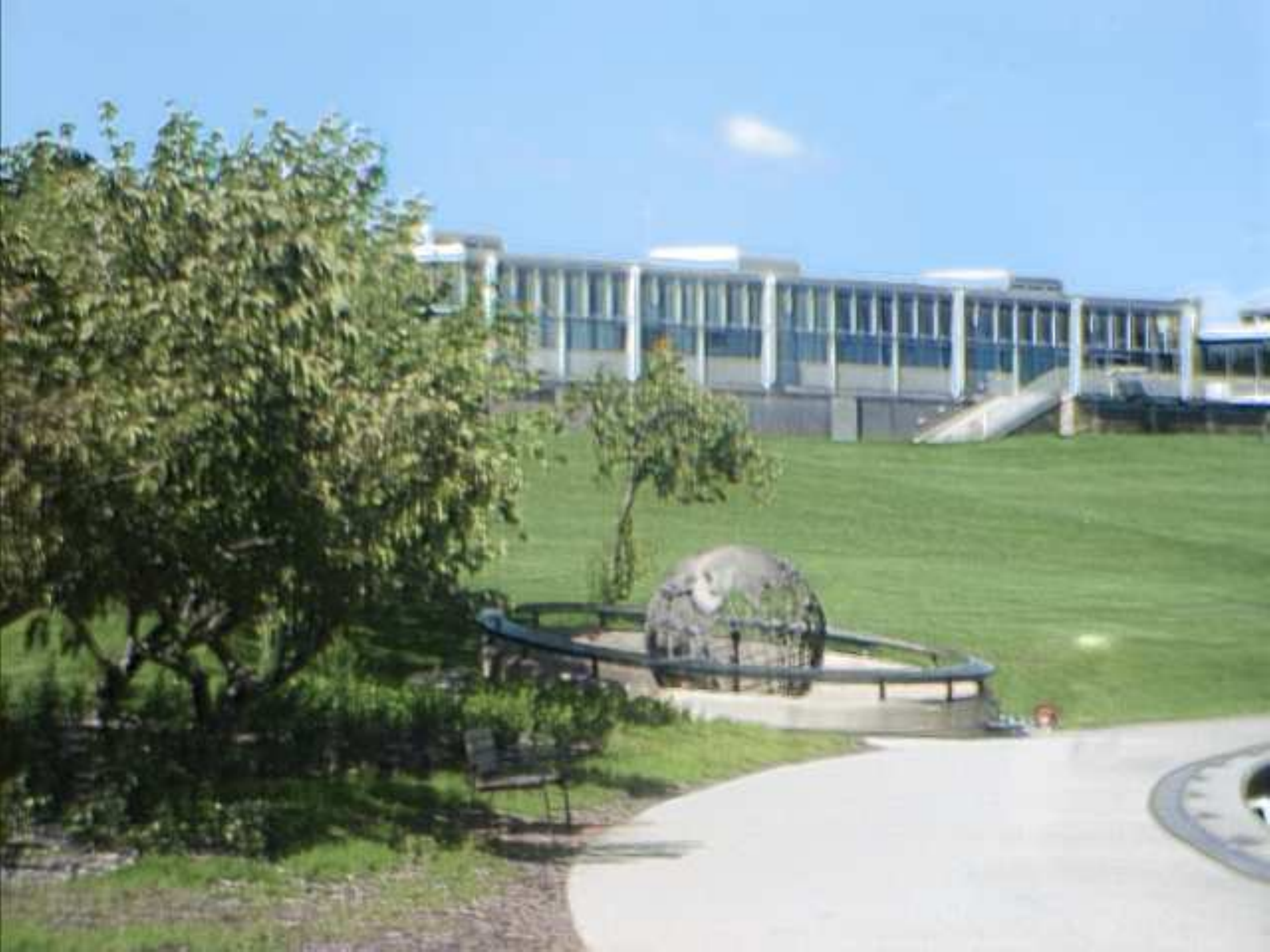}
\label{d04manifoldalignment}
}
\caption{Visual comparison of different visualization approaches on the D04 data set.}
\label{D04comparison}
\end{figure}
\begin{figure}[t]
\centering
\subfigure[]{
\includegraphics[width=0.3\linewidth]{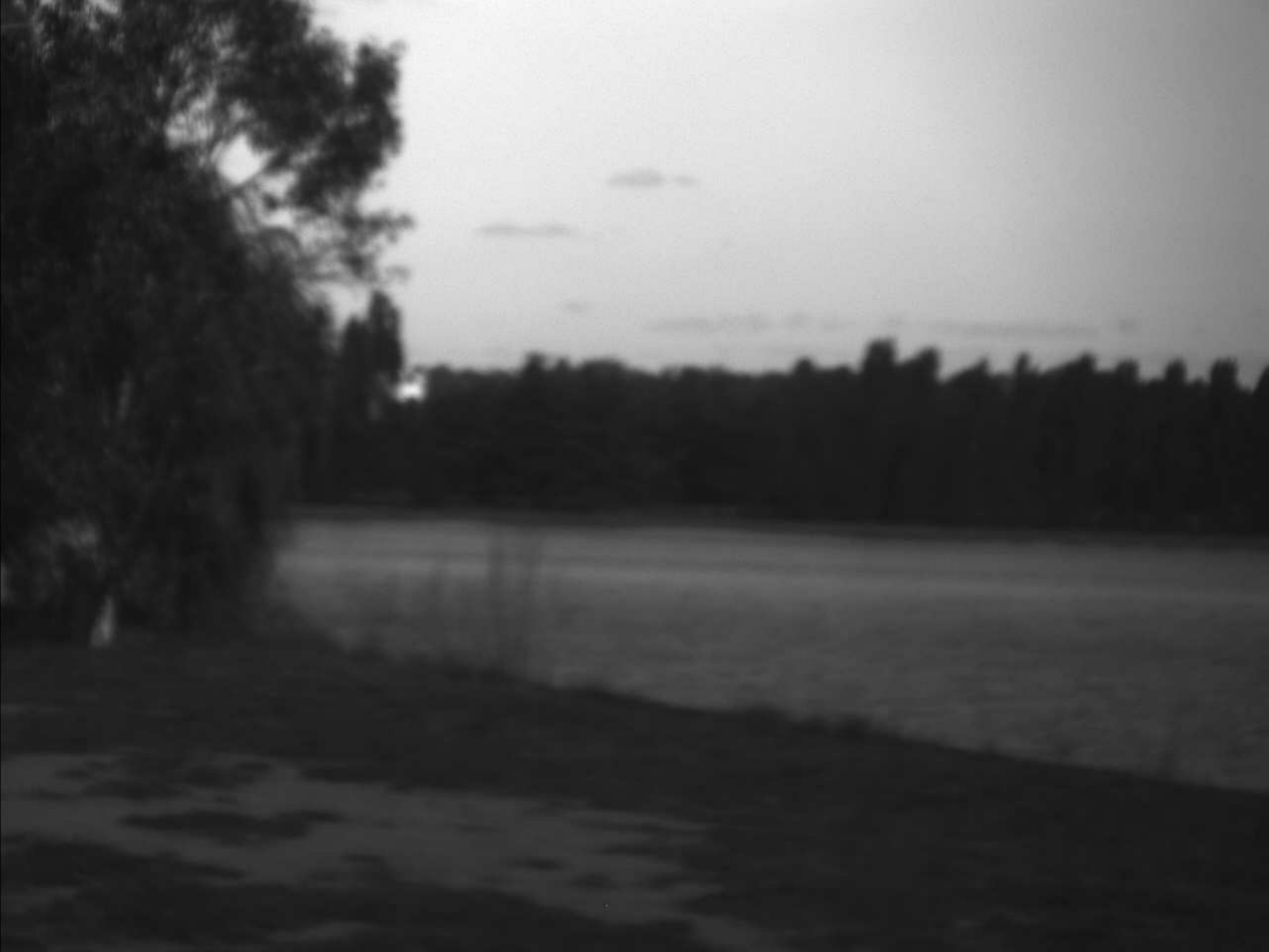}
}
\subfigure[]{
\includegraphics[width=0.3\linewidth]{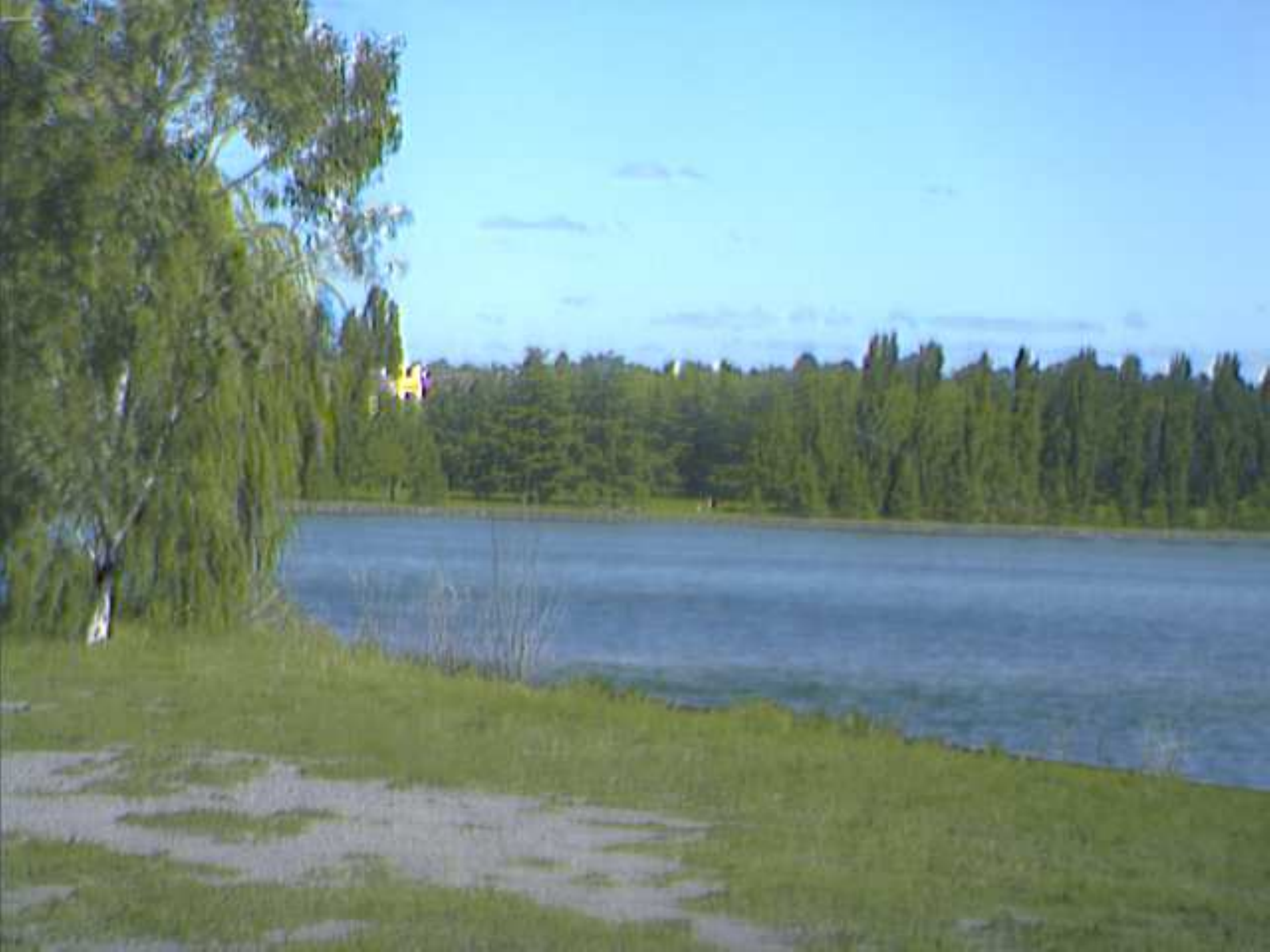}
}
\subfigure[]{
\includegraphics[width=0.3\linewidth]{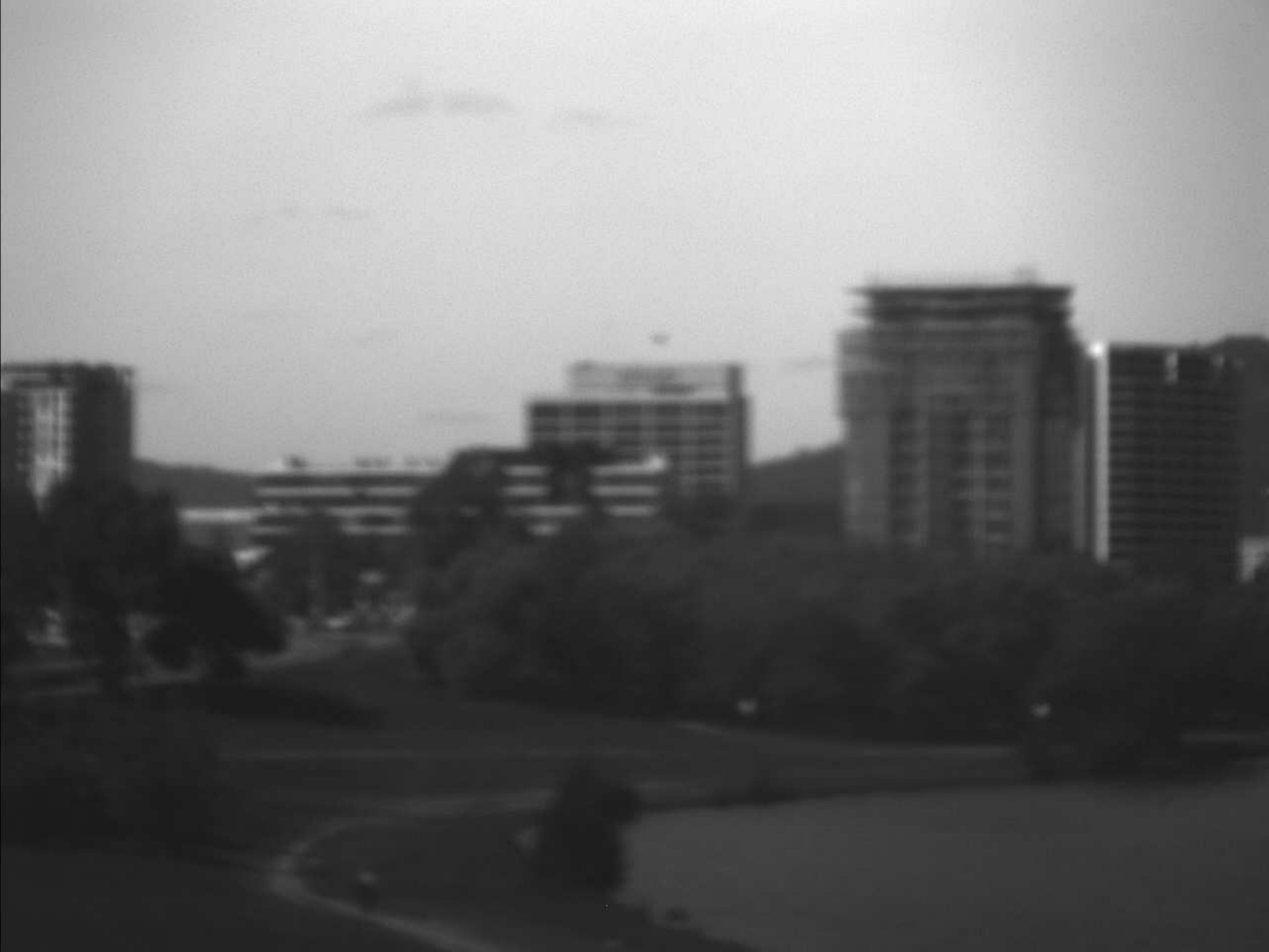}
}
\subfigure[]{
\includegraphics[width=0.3\linewidth]{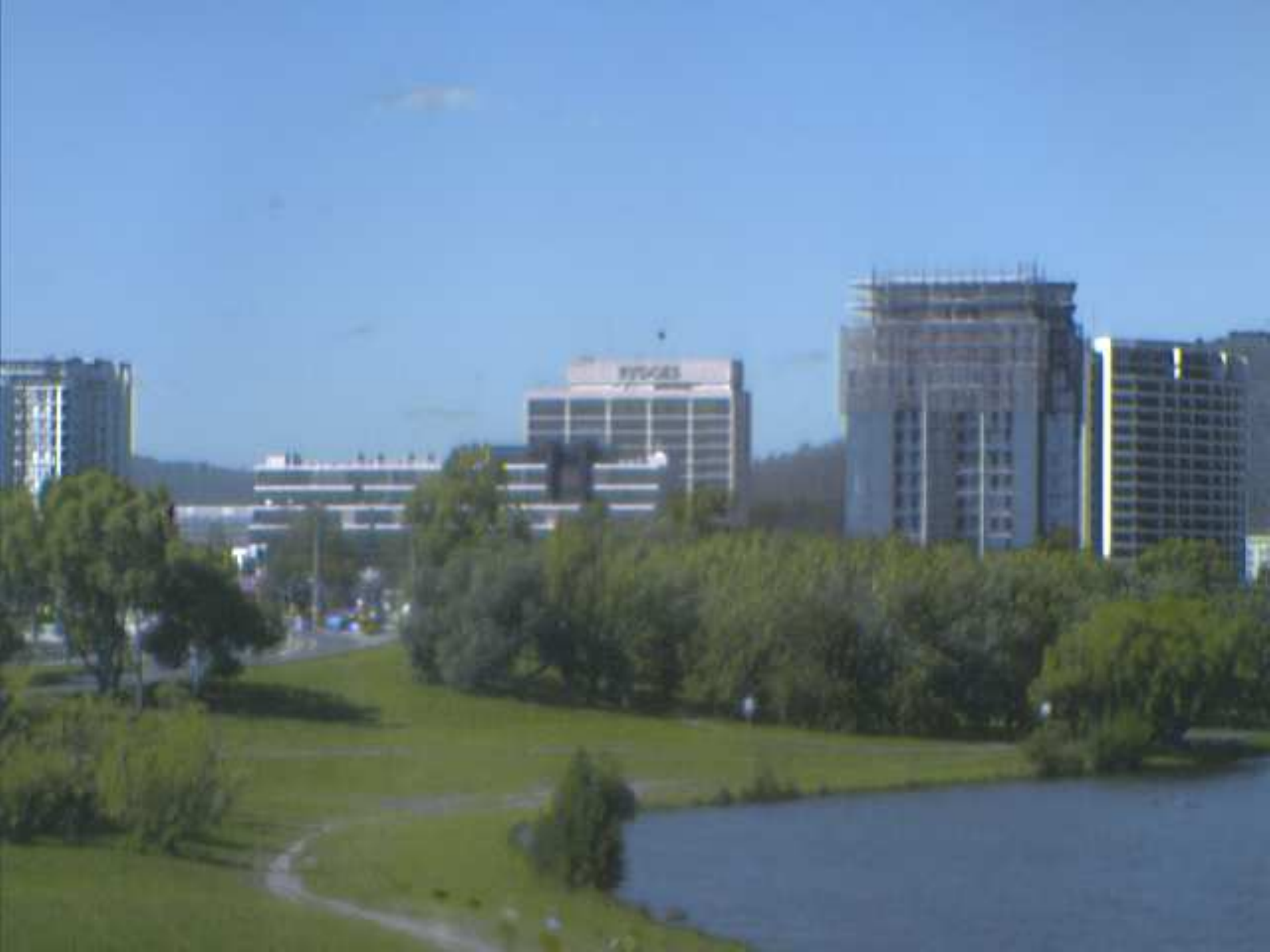}
}
\caption{Visualizations of similar HSIs generated by reusing a set of matching pixels.
(a) The first band of F02. (b) The visualization of F02. (c) The first band of G02. (d) The visualization of G02.}
\label{Generalize}
\end{figure}
\begin{table}[t]
\centering
 \caption{Performance comparison on entropy.}
  \label{entropytable}
  \begin{tabular}{ccccc}
    \hline
      Visualization methods& Washington& Moffett & G03& D04\\
     \hline
     LP Band Selection & 6.59 &6.36&4.73&5.58\\
    \hline
    Manifold Alignment & \bftab 7.40 & \bftab 6.89 & 6.64 & $7.02$\\
    \hline
    Stretched CMF & $3.89$ & $5.28$ & $6.31$ &  7.13\\
    \hline
    Bilateral Filtering&3.31 &5.87&6.05&  7.27\\
    \hline
    Bicriteria Optimization&5.75&6.00&5.73&6.06\\
    \hline
    MLS&7.00 &6.30& \bftab 7.01&\bftab 7.39\\
    \hline
  \end{tabular}
\end{table}
 \begin{table}
\centering
\caption{Performance comparison on RMSE.}
  \label{Root mean square error}
  \begin{tabular}{ccccc}
    \hline
      Visualization methods& Washington& Moffett & G03& D04\\
     \hline
    LP Band Selection& 68.35&49.93&95.19&83.35\\
        \hline
    Manifold Alignment &  58.41 &  27.30 &  32.88 & \bftab 23.08\\
    \hline
    Stretched CMF & $92.46$ & $58.50$ & $44.09$ & $28.70$\\
    \hline
    Bilateral Filtering& 102.43&56.70&78.54&61.14\\
    \hline
    Bicriteria Optimization&98.47&57.94&69.60&75.70\\
      \hline
     MLS& \bftab 46.85&\bftab 19.07&\bftab 16.45& 29.78\\
    \hline
  \end{tabular}
  \end{table}
\begin{figure}[t]
\centering
\subfigure[]{
\includegraphics[width=0.3\linewidth]{img/d04_band1.pdf}
\label{fig:D04band1_ds}
}
\subfigure[]{
\includegraphics[width=0.3\linewidth]{img/registered_g03.pdf}
\label{fig:G03RGB_ds}
}\\
\subfigure[]{
\includegraphics[width=0.3\linewidth]{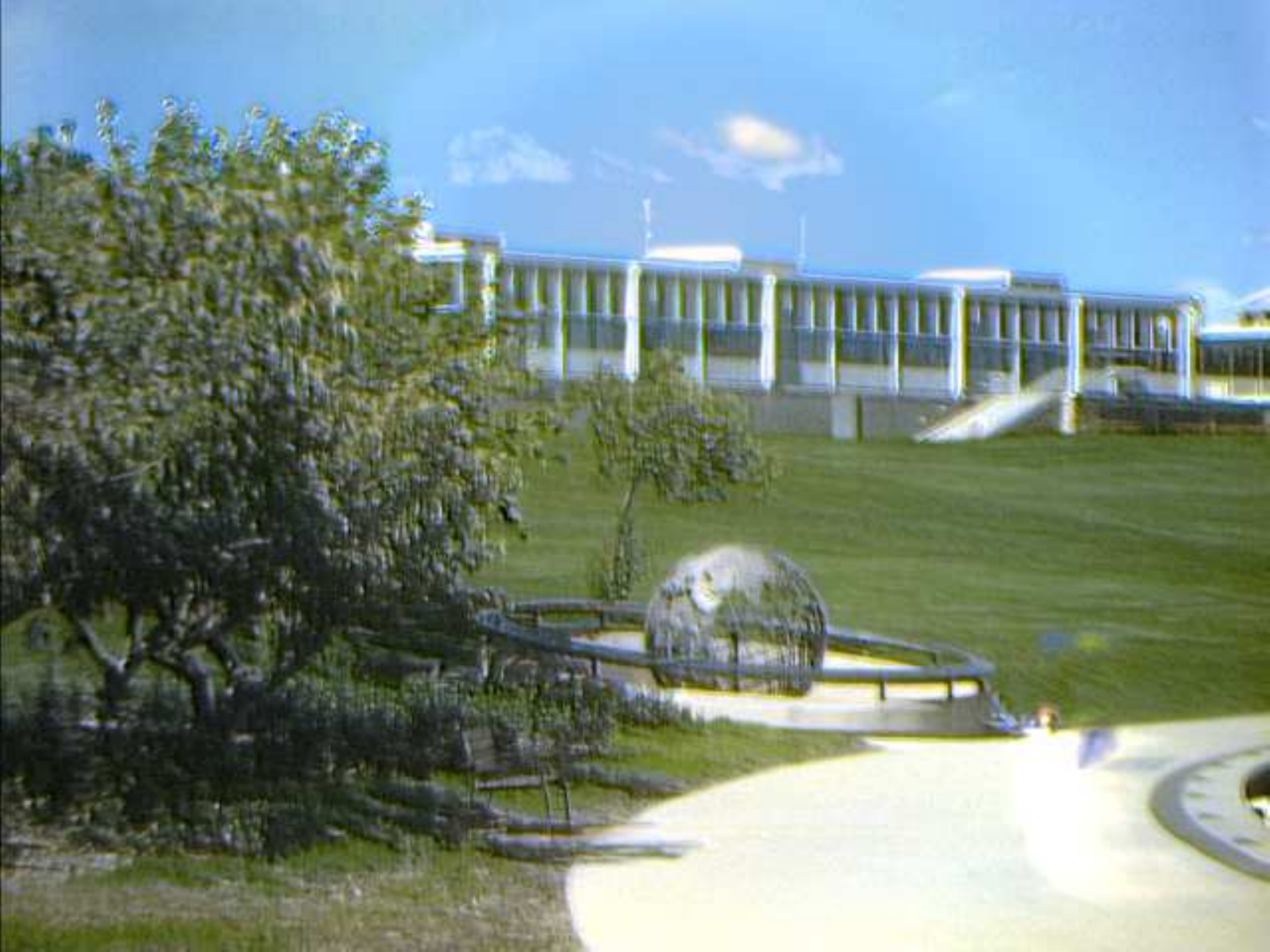}
\label{fig:d04andg03}
}
\caption{Visualization of D04 generated with the aid of an RGB image
captured on a different site. (a) The first band of D04. (b) An RGB image of different site. (c) The visualization of D04.}
\label{differentSites}
\end{figure}
\section{Experiments and Discussions}
\label{Experiments}
We compare our results against several recently
proposed visualization approaches including band selection using LP~\cite{su2014hyperspectral}, stretched CMF~\cite{jacobson2005design},
bilateral filtering~\cite{kotwal2010visualization}, bicriteria optimization~\cite{mignotte2012bicriteria} and manifold alignment~\cite{liaomanifold}.
We also give the experimental results on two other application scenarios.
\subsection{Hyperspectral Imaging Data}
Four HSI data sets were used to evaluate the performance of different visualization methods.
One remote sensing HSI data set was taken over the Washington D.C. mall by the Hyperspectral Digital Imagery Collection Experiment (HYDICE) sensor.
The data consists of 191 bands after noisy bands removed. Each band image has a size of $1208\times 307$.
Fig.~\ref{fig:band50} shows its 50th band image.
Fig.~\ref{fig:registered} shows the registered corresponding RGB image obtained from Google Earth.
Another remote sensing HSI data was captured by the Airborne Visible/Infrared Imaging Spectrometer (AVIRIS) over Moffett Field, California. The data consists of 224 bands, and each band has a size of $501\times 651$.
Its 50th band image is shown in Fig.~\ref{fig:MoffettBand50}.
The corresponding RGB image shown in Fig.~\ref{fig:RGB_moffett} was also obtained from Google Earth.

The other two HSI data sets named ``G03'' and ``D04'' were captured by a ground-based OKSI hyperspectral imaging system mounted with a tunable LCTF filter.
Each data set has 18 bands ranging from $460nm$ to $630nm$ at $10nm$ interval.
The size of each band is $960\times 1280$.
Fig.~\ref{fig:D04band1} and Fig.~\ref{fig:G03band1} show the first band images of D04 and G03 respectively.
Their corresponding RGB images were taken by a Nikon D60 SLR digital camera and are shown in Fig.~\ref{fig:RGB_D04} and Fig.~\ref{fig:RGB_G03} respectively.
\subsection{Comparison of Visualization Methods}
Fig.~\ref{DCcomparison} shows the visual comparison of different visualization approaches on the Washington D.C. mall data.
It can be seen that output image of the proposed method is very easy to understand
since it has very natural colors
similar to that of the corresponding RGB image in Fig.~\ref{fig:registered}.
Also, the details in the HSI are clearly represented.
The comparative results on the Moffett field, G03 and D04 data sets are provided in Fig.~\ref{MoffettExperiments}-\ref{D04comparison} respectively.
Likewise, in these experiments our approach not only shows great performance on displaying the HSIs with natural colors but also preserves the local structure of the HSIs.

Quantitative assessment of HSI visualization does not have a universally accepted standard.
In this paper, we adopt two quantitative metrics: entropy of the output image~\cite{kotwal2010visualization} and root mean square error (RMSE) between the output image and the corresponding RGB image~\cite{zhu2007evaluation}.

Entropy is a statistical measure of randomness that can be used to characterize the texture of an image.
An image with higher entropy generally contains richer information.
Table~\ref{entropytable} shows the comparative results in terms of entropy.
It can be found that the results of the proposed method have relatively large entropy compared to other approaches, which suggests that the proposed method is able to preserve the information of the HSI for analysis.

The RMSE between the output image and the corresponding RGB image is a straightforward way to evaluate the ``naturalness'' of the output image is~\cite{zhu2007evaluation}.
The comparative RMSE results are given in Table~\ref{Root mean square error}.
We can see that the output images of the proposed method have smaller RMSE compared to other methods in most cases, which shows clearly that the proposed method is an excellent approach for displaying HSIs with natural colors.
\subsection{Generalizing the Matching Pixels to Visualize Other HSIs}
Once we have obtained the matching pixels between an HSI and a corresponding RGB image, the matching pixels can be reused to visualize semantically similar HSIs captured by the same hyperspectral imaging sensor.
In this experiment, the matching pixels between
the G03 data set and its corresponding RGB image are applied to
visualize three HSI data sets named ``F02'' and ``G02'', which were captured by the same imaging sensor.
Fig.~\ref{Generalize} shows the first band images and the visualizations of these HSIs.
It can be seen that the proposed method has nice performance in displaying these HSIs with natural colors.
Besides, the same classes of objects are presented with consistent colors across the three images.
\subsection{Displaying an HSI with an RGB Image of a Different Site}
In this experiment, we visualize the D04 data set (see Fig.~\ref{fig:D04band1_ds}) aided by an RGB image captured on a different site (see Fig.~\ref{fig:G03RGB_ds}).
Fourteen matching pixel pairs between the two images were manually selected to serve as control points.
Fig.~\ref{fig:d04andg03} shows the color representation of D04.
We can see that this image has natural colors and maintains the basic information in the HSI, which shows that our algorithm performs well with a limited number of matching pixels between images captured on different sites.
\section{Conclusions}
\label{Conclusions}
We have presented a novel approach to visualize HSIs with natural colors based on the MLS interpolation scheme.
The key idea is to construct a 3D color representation of an input HSI through a set of matching pixels between the HSI and a corresponding RGB image.
Our method solves for each spectral signature a unique transformation so that the nonlinear
structure of the HSI can be preserved.
One advantage of the proposed method is that the matching pixels between an HSI and an RGB image can be directly applied to visualize similar HSIs captured by the same imaging sensor.
Our method also works well when the HSI and the corresponding RGB image are captured on
different sites.
In the future, we plan to take the spatial information into account to better preserve the spatial structure of HSIs.

\bibliographystyle{IEEEtran}
\bibliography{HSIvisualization}
\end{document}